\documentclass[twoside,11pt]{article}

\usepackage[abbrvbib,preprint]{dmlr2e} %

\usepackage[utf8]{inputenc} %
\usepackage[T1]{fontenc}    %
\usepackage{hyperref}       %
\usepackage{url}            %
\usepackage{booktabs}       %
\usepackage{amsfonts}       %
\usepackage{microtype}      %

\usepackage[svgnames]{xcolor}
\usepackage{array}
\usepackage{multirow}
\usepackage{makecell}
\usepackage{graphicx}
\usepackage{placeins}
\usepackage{enumitem}
\usepackage{csquotes}
\usepackage{wrapfig}
\usepackage{subcaption}
\usepackage{bm}
\usepackage[flushleft]{threeparttable}
\usepackage{mathtools}

\usepackage{colortbl}
\newcommand{\subtleCmidrule}[1]{%
    \arrayrulecolor{gray}%
    #1%
    \arrayrulecolor{black}%
}

\usepackage{cleveref}
\crefname{subfigure}{Subfigure}{Subfigures}
\Crefname{subfigure}{Subfigure}{Subfigures}
\AddToHook{cmd/appendix/before}{\crefalias{section}{appendix}}
\AddToHook{cmd/appendix/before}{\crefalias{subsection}{subappendix}}
\AddToHook{cmd/appendix/before}{\crefalias{subsubsection}{subsubappendix}}

\usepackage{pifont}
\newcommand{\cmark}{\textcolor{ForestGreen}{\ding{51}}}
\newcommand{\xmark}{\textcolor{DarkRed}{\ding{55}}}

\newcommand{\denseColumns}[0]{\setlength{\tabcolsep}{5.1pt}}

\newcommand{\R}[0]{\mathbb{R}}
\DeclareMathOperator{\Var}{Var}

\usepackage{listings}
\definecolor{lightgray}{gray}{0.95}
\usepackage{tcolorbox}
\newtcolorbox{llmprompt}[1]{%
  colback=blue!5, colframe=blue!30, boxrule=1.5pt, width=\textwidth, 
  title=\vspace{3pt}#1, fonttitle=\bfseries,
  arc=2mm,
  fontupper=\small
}

\newcommand{\datasetName}{QuAnTS}
\newcommand{\ourTitle}{\datasetName{}: Question Answering on Time Series}

\newcommand{\authorLastNames}{Divo, Kraus, Nguyen, Xue, Razzak, Salim, Kersting, and Dhami}
\newcommand{\linkGeneration}[0]{\url{https://github.com/mauricekraus/quants-generate}}
\newcommand{\linkEval}[0]{\url{https://github.com/felixdivo/quants-eval}}
\newcommand{\linkDataset}[0]{\url{https://huggingface.co/datasets/dasyd/quants}}
\newcommand{\doiDataset}[0]{\href{https://doi.org/10.57967/hf/6663}{10.57967/hf/6663}}

\dmlrheading{1}{2026}{1-\pageref{LastPage}}{11/25; Revised 08/26}{10/26}{21-0000}{\authorLastNames{}}
\def\openreview{\url{https://openreview.net/forum?id=bNCDElSOXB}}

\ShortHeadings{\ourTitle}{\authorLastNames{}}

\begin{document}

\title{\ourTitle}

\author{\name Felix Divo\normalfont{\textsuperscript{1,*}} \email felix.divo@cs.tu-darmstadt.de
    \AND
    \name Maurice Kraus\normalfont{\textsuperscript{1,*}} \email maurice.kraus@cs.tu-darmstadt.de
    \AND
    \name Anh Q. Nguyen\normalfont{\textsuperscript{2}} \email anhq.nguyen23@gmail.com
    \AND
    \name Hao Xue\normalfont{\textsuperscript{3}} \email haoxue@hkust-gz.edu.cn
    \AND
    \name Imran Razzak\normalfont{\textsuperscript{4}} \email imran.razzak@mbzuai.ac.ae
    \AND
    \name Flora D. Salim\normalfont{\textsuperscript{2}} \email flora.salim@unsw.edu.au
    \AND
    \name Kristian Kersting\normalfont{\textsuperscript{1,5,6,7}} \email kersting@cs.tu-darmstadt.de
    \AND
    \name Devendra Singh Dhami\normalfont{\textsuperscript{8}} \email d.s.dhami@tue.nl
    \AND
    \normalfont{\footnotesize
        \textsuperscript{1}AIML Lab, TU Darmstadt\\
        \textsuperscript{2}University of New South Wales\\
        \textsuperscript{3}Hong Kong University of Science and Technology (Guangzhou)\\
        \textsuperscript{4}MBZUAI, Abu Dhabi\\
        \textsuperscript{5}Hessian Center for AI~(hessian.AI)\\
        \textsuperscript{6}German Research Center for AI~(DFKI), Darmstadt\\
        \textsuperscript{7}Centre for Cognitive Science, TU Darmstadt\\
        \textsuperscript{8}TU Eindhoven\\
        \textsuperscript{*}Authors contributed equally\
    }
}

\editor{Hongyang Zhang}

\maketitle
\label{first_page}

\begin{abstract}
    Text offers intuitive access to information.
    This can, in particular, complement the density of numerical time series, thereby enabling improved interactions with time series models to enhance accessibility and decision-making.
    While the creation of question-answering datasets and models has recently seen remarkable growth, most research focuses on question answering~(QA) on vision and text, with time series receiving minute attention.
    To bridge this gap, we propose a challenging novel time series QA~(TSQA) dataset, \datasetName{}, for \textbf{Qu}estion \textbf{An}swering on \textbf{T}ime \textbf{S}eries data.
    Specifically, we pose a wide variety of questions and answers about human motion in the form of tracked skeleton trajectories.
    We verify that the large-scale \datasetName{} dataset is well-formed and comprehensive through extensive experiments.
    Thoroughly evaluating existing and newly proposed baselines then lays the groundwork for a deeper exploration of TSQA using \datasetName{}.
    Additionally, we provide human performances as a key reference for gauging the practical usability of such models.
    We hope to encourage future research on interacting with time series models through text, enabling better decision-making and more transparent systems.
\end{abstract}


\clearpage

\section{Introduction}
\label{sec:intro}

\begin{wrapfigure}{r}{0.4\textwidth}
    \centering
    \vspace{-\baselineskip}
    \includegraphics[width=\linewidth]{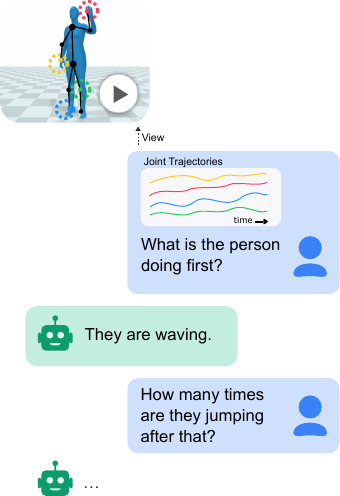}
    \caption{\textbf{Text offers a highly intuitive medium of communication to interact with otherwise opaque multivariate time series.}}
    \label{fig:motivational_chat}
    \vspace{-\baselineskip}
\end{wrapfigure}

Language is a highly intuitive medium of communication~\citep{fedorenkoLanguagePrimarilyTool2024}.
Humans learn to listen and verbally express themselves early in life, and most kids learn to read and write soon after.
For example, we can easily describe perceived objects, their relations, chains of events, theories, uncertainties, and much more in this unifying modality~\citep{walkerToddlersInferHigherOrder2014,jiangMEWLFewshotMultimodal2023}.
However, in the age of automation, we often encounter other kinds of data, including temporal data consisting of numerical values recorded over time.
Time series are complex and precise pieces of data that are highly relevant in many applications.
In contrast to the natural use of text, however, end users are often not immediately able to understand and interact with them~\citep{eggletonIntuitiveTimeSeriesExtrapolation1982} and might require specific training and tooling~\citep{haroldMakingSenseTimeSeries2015,zhaoInterpretationTimeSeriesDeep2023}.
Therefore, employing text as a flexible bridge to more intuitively access time series is an exciting opportunity~\citep{imaniPuttingHumanTime2019,slackExplainingMachineLearning2023}.
\Cref{fig:motivational_chat} shows how this even allows for back-and-forth discussions contextualized by data.

Practical machine learning applications require solving a wide variety of tasks on time series, including the classics of forecasting, classification, and anomaly detection.
While tremendous progress has been made in building models for each individual use, most are only effective at one, and changes necessitate expensive remodeling and retraining.
In the era of large language models~(LLMs), we can solve many tasks by embedding both our data and the assignment description as text~\citep{brownLanguageModelsAre2020,lewisRetrievalAugmentedGenerationKnowledgeIntensive2020,fanSurveyRAGMeeting2024}.
This often does not require specific retraining, as providing some examples by in-context learning is often sufficient.
It would be highly desirable to tweak outputs of time series models in a similarly flexible way, as some initial works already started exploring~\citep{xieChatTSAligningTime2025}.
However, interacting with time series via text is only just emerging as a field of study.

This combination is further motivated by the common need to enrich data with further context and task descriptions, which is eased by the possibility of encoding such information in text~\citep{hanECOLAEnhancingTemporal2023}.
One example is multimodal forecasting, where textual data enriches numerical time series to provide crucial context~\citep{jiaGPT4MTSPromptbasedLarge2024,williamsContextKeyBenchmark2025}.
Interactive feedback promises even better results by allowing a human-in-the-loop to offer richer information~\citep{imaniPuttingHumanTime2019,wuSurveyHumanintheloopMachine2022,kraus2025right}.
Generally, time series are used to describe and subsequently analyze weather dynamics~\citep{hewageTemporalConvolutionalNeural2020}, plant growth~\citep{yasrabPredictingPlantGrowth2021}, electrocardiograms~\citep{sahooMachineLearningApproach2020}, electrical brain activity~\citep{andrzejakIndicationsNonlinearDeterministic2001}, human activities~\citep{laraSurveyHumanActivity2013}, manufacturing processes~\citep{farahaniTimeseriesPatternRecognition2023}, traffic volumes~\citep{jiangGraphNeuralNetwork2022}, financial markets~\citep{nazarethFinancialApplicationsMachine2023,divoForecastingCompanyFundamentals2025}, and much more.
In all those fields, combining time series with text could help domain experts better leverage data to drive more informed decisions.
For instance, one could supplement traffic volume predictions with information about local festivities, or human activity recognition with details on a person's disabilities.
In other tasks, the textual domain not only serves as context but is an integral part of it, such as answering questions about time series or their textual summarization~\citep{novakLinguisticCharacterizationTime2016}.

Yet, we find that very little work has been performed in that direction.
We attribute this mainly to insufficient high-quality data.
This limits both the training data to learn from and the opportunities to evaluate new systems.
Therefore, we present \textbf{\datasetName{}}: A challenging multimodal dataset for question answering on numerical time series data.
We use human motion, in the form of tracked skeleton positions, as the signal to reason about.
This has several key advantages:
(1)~It is natural to visualize the numerical values as human figures, making them interpretable for humans while still challenging for machines.
(2)~This field of application is central to building intelligent devices for, e.g., applications in health and sports.
(3)~There is a very long line of research on and collection of human motion data, highlighting its significance and providing tools and insights we can leverage in this work, such as the synthetic motion generation and visualization tools.

To move toward rich interactions with complex time series signals via textual conversations, we \textbf{contribute} the following:
\begin{enumerate}[label=(\roman*),topsep=2pt,itemsep=0pt]
    \item We investigate the design space of time series QA datasets to compile a comprehensive and challenging set of question and answer types.
    \item We propose a procedure for generating the diverse and large-scale \datasetName{} dataset. The implementation is readily extensible and freely available alongside the dataset.\footnote{%
        We provide the code for generating \datasetName{} at \linkGeneration{} and for evaluation at \linkEval{}.
        The dataset is available from \linkDataset{} with DOI \doiDataset{}.
    }
    \item Extensive dataset validation and benchmarking of existing and novel baseline models provide a fertile ground for further research into this very young field.
    \item We provide human performance to contextualize current and future machine capabilities.
\end{enumerate}

\paragraph{Outline.}
We continue by reviewing related work in \Cref{sec:related_work} as a foundation for the subsequent description of \datasetName{}'s construction in \Cref{sec:dataset}.
Extensive empirical findings on solving \datasetName{} are then presented in \Cref{sec:bench}.
We conclude the work with limitations and an outlook in \Cref{sec:conclusion} and provide an overview of its broader impact.

\section{Related Work}
\label{sec:related_work}

Integrating data sources and models from multiple modalities, such as images, videos, text, time series, tables, graphs, point clouds, etc., has experienced tremendous interest and progress in the last few years.
For instance, vision-language models have since arrived in mass usage as new commodity services~\citep{rombachHighResolutionImageSynthesis2022,openaiGPT4TechnicalReport2024a,zongSelfSupervisedMultimodalLearning2024,wadekarEvolutionMultimodalModel2024}.
Initial attempts have since worked toward any-to-any multimodal learning, i.e., aligning representations across more than two modalities \citep{hanOneLLMOneFramework2024,zhangMetaTransformerUnifiedFramework2023,wuNExTGPTAnytoAnyMultimodal2024}.
However, this field is still emerging and far from mature.

\subsection{Multimodal Time Series Models and Datasets}
\label{sec:related_work:mmts}

Fusing time series with other modalities remains largely unexplored, with a significant portion of multimodal research focusing on language and vision~\citep{baltrusaitisMultimodalMachineLearning2019,gaoSurveyDeepLearning2020}.
This is reflected and possibly caused by a shortage of truly multimodal time series datasets~\citep{liangMultiZooMultiBenchStandardized2023,zhouMoTimeDatasetSuite2025}, which is changing only very recently.

Some works already combined time series and text, particularly for augmenting tasks such as forecasting, classification, and anomaly detection with supplemental text data~\citep{sanoMultimodalAmbulatorySleep2019,dangSquawkBotJoint2020,jiaGPT4MTSPromptbasedLarge2024,liuTimeMMDMultiDomainMultimodal2024,chanMedTsLLMLeveragingLLMs2024,jinTimeLLMTimeSeries2024,kohautConstitutionalFilterBayesian2025,kohautDrivingFastSlow2026}.
However, in such tasks, the text often serves only as additional context, rather than being integral to the problem.
Thus, newer works focus on the text being essential for solving the tasks~\citep{williamsContextKeyBenchmark2025,xuInterventionAwareForecastingBreaking2025}.

These tasks, however, are solved by predicting time series snippets or labels, but not textual responses, such as in time series captioning~\citep{spreaficoNeuralDataDrivenCaptioning2020,imranLLaSALargeMultimodal2024,hanOneLLMOneFramework2024,trabelsiTimeSeriesLanguage2025,stremCATSCrossModalAutoencoding2026}.
These methods, however, do not adapt their responses based on questions or other textual input.
This is the case in QA over Temporal Knowledge Graphs~(TKGs) \citep{saxenaQuestionAnsweringTemporal2021,suTemporalKnowledgeGraph2024}, where events and their semantic relations have already been identified, or temporal QA, where questions with a specific focus on temporal reasoning based on purely textual context are to be answered~\citep{jiaTempQuestionsBenchmarkTemporal2018,jinForecastQAQuestionAnswering2021}. %
While reasoning over TKGs has since been extended to QA on forecasting~\citep{dingForecastTKGQuestionsBenchmarkTemporal2023}, it is not tied to the general context of a numerical time series and thus different from the approach of \datasetName{}.

Some initial works have already begun exploring true time series QA~(TSQA) and provided datasets, as summarized in \Cref{tab:dataset-stats-comparison}.
Firstly, \citet{xingDeepSQAUnderstandingSensor2021} propose the closed-ended OppQA dataset based on existing human activity recordings and the DeepSQA set of baselines.
Similarly, \citet{ohECGQAComprehensiveQuestion2023} annotated electrocardiogram recordings to obtain the EGC-QA dataset.
However, responses are again framed as a classification problem of one or multiple predefined answers.
The Time Series Evol-Instruct dataset accompanying the ChatTS model~\citep{xieChatTSAligningTime2025} provides open-ended QA data across various fields, time series lengths, and numbers of variates.
Lastly, \citet{imranLLaSALargeMultimodal2024} propose (Tune\nobreakdash-)OpenSQA, a human activity TSQA dataset again featuring exclusively open-ended questions.
It was obtained by prompting an LLM to generate temporal questions. 
They focus on a single activity, thereby lacking the completeness of downstream model training and evaluation on the entirety of the scene.
Moreover, the diversity of (Tune\nobreakdash-)OpenSQA is bounded by the diversity that emerges from a single prompt and an LLM.
\datasetName{} expands on these works by simultaneously featuring binary/multiple-choice and open-ended responses on a substantial scale and with assured diversity.
Furthermore, we ask questions about sequences of consecutive actions and their composition, necessitating models that can both perceive and reason over time series.
In addition, the human motion data in \datasetName{} challenges models by offering substantially more variates than any other dataset mentioned above~(apart from the substantially less diverse OppQA).
To contextualize the progress of the emerging field of TSQA, we provide human reference performances on \datasetName{} for the first time.

\begin{table}[t]
    \centering
    \caption{\textbf{The \datasetName{} dataset is large and varied.}
        It guarantees diversity through the balanced distribution of 45 question and three answer types: \emph{b}inary, \emph{m}ultiple-choice, and \emph{o}pen.
        The shape of the time series is specified in terms of the number of time steps $T$ and variates $V$.
        \datasetName{} provides human performances for reference.}
    \label{tab:dataset-stats-comparison}
    \scriptsize
    \setlength{\tabcolsep}{2pt}
    \begin{threeparttable}
    \begin{tabular}{@{\hspace{1pt}}lcccccccc@{\hspace{2pt}}c@{\hspace{2pt}}cccc@{\hspace{1pt}}}
        \toprule
        & & & \multicolumn{2}{c}{QA Pairs} & \multicolumn{3}{c}{Questions} & \multicolumn{3}{c}{Answ.} & \multicolumn{2}{c}{Time Series} & \\
        \cmidrule(lr){4-5} \cmidrule(lr){6-8} \cmidrule(lr){9-11} \cmidrule(lr){12-13}
        Dataset & Scope & Level & \#Total & \#Unique & \#Unique & Types & Dist. & b & m & o & \#Unique\hspace{-1ex} & \phantom{0,00}T$\times$V\phantom{0} & \multirow[t]{2}{*}{\shortstack{Human \\ Eval.}} \\
        \midrule
        OppQA-1800\tnote{1} & scene & high & 112,749 & \phantom{0}89,735 & \phantom{0}84,389 & \phantom{0}12 & balanced & \cmark & \cmark & \xmark & \phantom{0}1,362 & 1,800$\times$77 & \xmark \\
        ECG-QA & scene & high & 414,348 & 112,455 & \phantom{0}34,767 & \phantom{00}7 & balanced & \cmark & \cmark & \xmark & 16,052 & 5,000$\times$12 & \xmark \\
        TS Evol-Instr. & mixed & low & \phantom{0}44,802 & \phantom{0}44,516 & \phantom{0}43,973 & \phantom{0}10 & skewed & \xmark & \xmark & \cmark & \phantom{0}4,127 & \phantom{00}varying & \xmark \\
        (Tune-)OpenSQA & local & low & 183,304 & 172,379 & 183,042 & n/a & skewed & \xmark & \xmark & \cmark & 35,683 & \phantom{00}varying & \xmark \\
        \textbf{\datasetName{} (ours)} & scene & high & 150,000 & 125,726 & 100,568 & \phantom{0}45 & balanced & \cmark & \cmark & \cmark & 30,000 & \phantom{0,}320$\times$72 & \cmark \\
        \bottomrule
    \end{tabular}
    \vspace{2pt}
    \begin{tablenotes}
        \tiny \item[1] OppQA-1800's \enquote{counting} answer type is multiple-choice since one picks from a fixed set of four answers.
      \end{tablenotes}
    \end{threeparttable}
\end{table}

\subsection{Human Motion Data and LLMs}
Human motion data, typically captured by wearable sensors such as accelerometers and gyroscopes, presents unique challenges due to its temporal complexity.
Since the rise of leveraging LLMs for general time series tasks, researchers have begun exploring their potential in processing and interpreting human motion data~\citep{xuePromptCastNewPromptBased2023,jinTimeLLMTimeSeries2024,jiaGPT4MTSPromptbasedLarge2024,zhangLargeLanguageModels2024}, despite some more recent questioning~\citep{jinPositionWhatCan2024,tanAreLanguageModels2025,zhangTextTimeRethinking2025,merrillLanguageModelsStill2024,jiangEmpoweringTimeSeries2024}.
This shift highlights the growing interest in adapting LLMs to handle motion datasets' complex temporal patterns and cross-variate characteristics, such as in zero-shot activity recognition based on inertial measurement units~\citep{jiHARGPTAreLLMs2024}.

The mainstream of current studies (such as \citet{zhou2023tent, moon2023imu2clip,liSensorLLMAligningLarge2025}) focuses on aligning sensor data embeddings with textual description embeddings extracted by LLMs.
However, a significant challenge in this approach arises from the limitations of existing human motion datasets, as most lack detailed textual descriptions of the activities performed.
In recent studies on LLMs for sensor data, the text component is often generated post-hoc through labor-intensive annotation processes, resulting in a lack of standardized criteria for them.
This inconsistency complicates the alignment of TS and text embeddings, particularly between different systems, making fair comparisons challenging.

This work presents the readily extensible, large-scale \datasetName{} dataset and generation pipeline comprising synthetic human motion data, corresponding textual descriptions, and challenging question-answer pairs.
It provides a new data source to support research on time series and textual embedding alignment while simultaneously posing a novel QA challenge.

\section{\datasetName{}: A Dataset for Question Answering on Time Series}
\label{sec:dataset}

\begin{figure}[t]
    \centering
    \includegraphics[width=\linewidth]{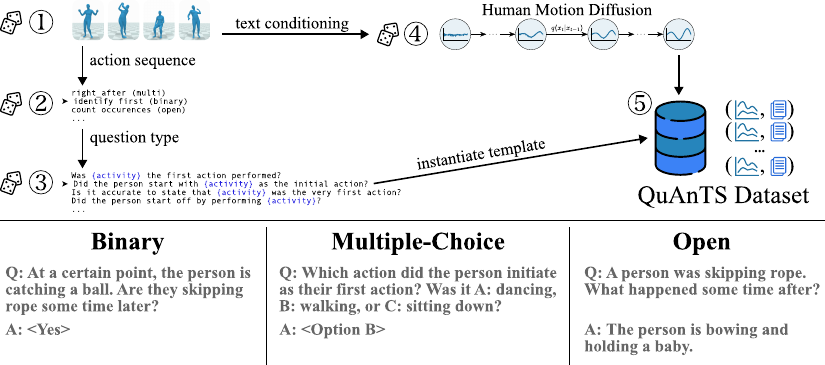}
    \caption{\datasetName{} is generated in several steps: 
        An action sequence is sampled \ding{192}, where for each we sample five question and answer types \ding{193}.
        For diversity, each of them is then instantiated from a sampled template \ding{194}.
        The time series from the human motion diffusion \ding{195} is then combined with the QA-pair and auxiliary data \ding{196}.
        Example QA pairs are shown below.
        Dice (\includegraphics[height=1.65ex]{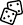}) indicate randomized operations to enhance dataset diversity.
    }
    \label{fig:overview}
    \vspace{-0.5\baselineskip}
\end{figure}

We created the diverse, large-scale \datasetName{} dataset as a foundation for training future time series QA models and evaluating them.
This section describes the design considerations taken into account during its development.
Firstly, we present how the textual descriptions with questions and answers were generated~(\Cref{sec:dataset:text}), followed by the text-conditioned generation of human motion trajectories~(\Cref{sec:dataset:ts}).
\Cref{fig:overview} shows the whole pipeline for generating \datasetName{}.

\subsection{Question \& Answer Generation Pipeline}
\label{sec:dataset:text}
\begin{figure}[t]
    \centering
    \includegraphics[width=0.9\linewidth]{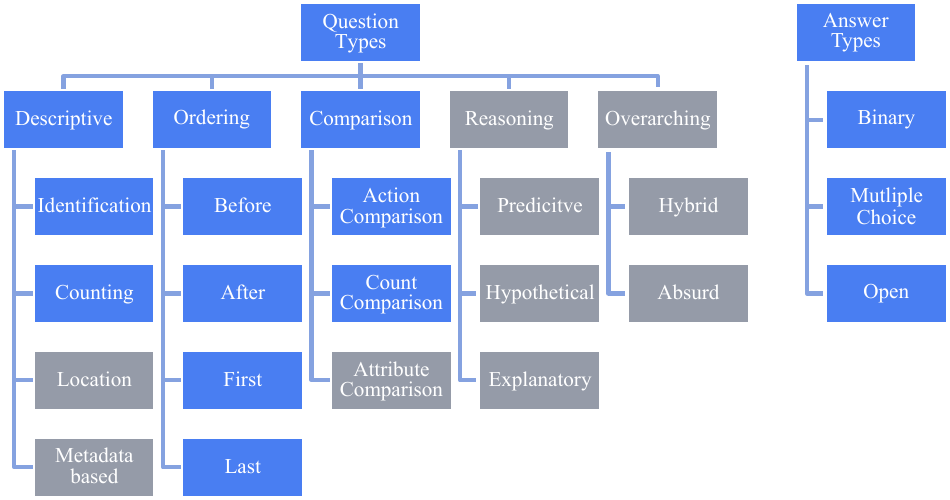}
    \caption{\textbf{The hierarchy of question and answer types we identified.}
        The \textcolor[HTML]{547AEE}{\textbf{blue}} types are the most fundamental to time series QA and, therefore, included in \datasetName{}.
        These tasks alone already make for a very challenging dataset, as we will see in \Cref{sec:bench}.
        The \textcolor[HTML]{969BA7}{\textbf{gray}} ones are left for future extensions.}
    \label{fig:question-types}
\end{figure}

Determining suitable questions and corresponding answers is crucial to obtaining a challenging dataset.
Question types can either be not categorized at all, specified beforehand, or assigned after manual dataset creation.
Since we generate questions synthetically by instantiating templates, it is natural to define them a priori to break down the data generation process into specific parts.
This has the added benefit of allowing for the inspection of a model's performance on specific subtasks and ensuring a balanced distribution of question types.
Following a substantial review of QA datasets and pipelines for time series and videos, we identified these five main question categories~(as shown in \Cref{fig:question-types}):
\begin{description}
    \setlength{\itemsep}{-0.3\baselineskip}
    \item[\textbf{Descriptive}] questions ask about the name of a performed action, how often it occurred, where it is happening, or insight based on metadata, such as who is performing the action. We exclude the latter two since we abstract from the concrete location, and metadata is often only available in lab settings.
    \item[\textbf{Temporal Ordering}] questions ask about the temporal ordering of events or if specific actions were performed in the very beginning or end.
    \item[\textbf{Comparison}] questions ask whether some actions are the same/distinct or if they occur the same/different number of times.
    \item[\textbf{Reasoning}] questions ask for higher-level concepts in the actions, like predicting the next one, counterfactual reflections, or providing reasons for a person's behavior. Once basic time series QA matures, we propose this as exciting future work.
    \item[\textbf{Overaching}] questions are a future type that either combines multiple aspects, like counting how often the last action was performed in total, or absurd questions, like asking about what happened after an action that never took place. We omit it for now, as \datasetName{} is already very challenging without it.
\end{description}

As in video QA, the simplest possible answer is a binary \enquote{Yes} or \enquote{No}.
\datasetName{} goes beyond such simple formats by offering multiple-choice and open-ended questions generated from templates, where we do not expect a label but a textual response.
Differentiating between these three answer formats, \datasetName{} comprises a balanced distribution of a total of 45 question types.
To ensure high diversity in \datasetName{}, we sample from about 10 to 20 different question and answer templates, depending on the QA type.
To increase data availability for training for open answers, we additionally generate textual answers for binary and multiple-choice questions via templating.
We leveraged LLM-assisted paraphrasing, followed by a comprehensive manual review, to enhance the diversity of all templates.

\subsection{Leveraging Motion Diffusion Models for Time Series Context}
\label{sec:dataset:ts}
Having generated textual descriptions of the desired sequence of actions and corresponding QA pairs, we next require matching time series.
We opted to use synthetic trajectories from a generative human motion model.
In contrast to directly using an existing dataset, this approach enables us to generate action sequences tailored to the specific questions we want answered.
Furthermore, the human motion generator can sample multiple variations of the same action, reflecting the variability of organic movements.
We require a method that allows us to generate motions smoothly following a sequence of activities, and not just a single one, to enable us to ask interesting questions about them.
Numerous generative human motion models can be conditioned on natural language prompts, much like diffusion-based vision models work.
Due to perceptually well-recognizable motions and very fast generation, we choose to use \emph{Spatio-Temporal Motion Collage}~(STMC) by \citet{petrovichMultiTrackTimelineControl2024}. %

To generate well-recognizable human motions, we identified 19 particularly distinct action prompts commonly found in STMC's training data~(see \Cref{sec:app:details-generate-dataset}) to build action sequences.
We evaluate their identifiability by humans in \Cref{sec:app:eval-details:human}.
Each context time series in our dataset was obtained by concatenating four action prompts of four seconds each from unique random seeds.
Five question-and-answer pairs share one context time series, ensuring high diversity at reasonable generation costs.
STMC first assigns textual prompts to one or multiple body parts to create a multi-track timeline.
To obtain the individual components, text-guided iterative denoising is performed for each track and time segment.
An additional refinement step is required to smoothly blend the transitions between consecutive sections with different prompts.
Specifically, we add back some unconditional noise to the transition region before a final denoising procedure, which is then conditioned by the already generated actions before and after the change point. %
This results in final poses comprising three spatial coefficients for each of the $24$ joints in the SMPL scheme~\citep{loperSMPLSkinnedMultiperson2015}.
The final trajectory of 16 seconds at 20 frames per second thus has dimension $320 \times 24 \times 3$.
\Cref{sec:app:details-generate-dataset} provides the full details of the generation procedure.

\subsection{Dataset Statistics and Diversity Analysis}
\label{sec:dataset:stats}

\begin{wrapfigure}{r}{0.41\textwidth}
    \centering
    \vspace{-\baselineskip}
    \includegraphics[width=0.95\linewidth]{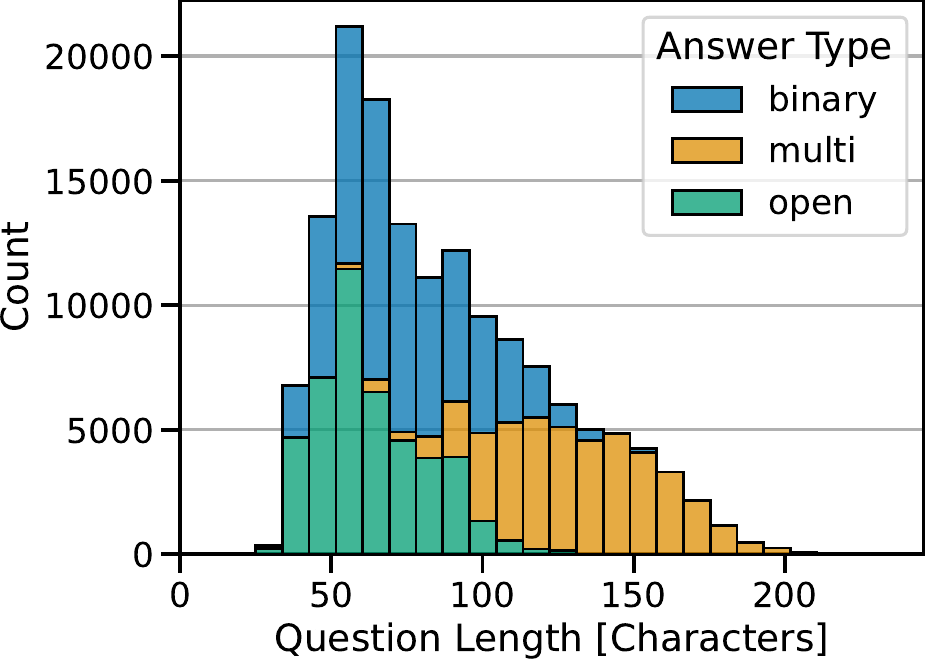}
    \vskip0.65\baselineskip
    \includegraphics[width=0.95\linewidth]{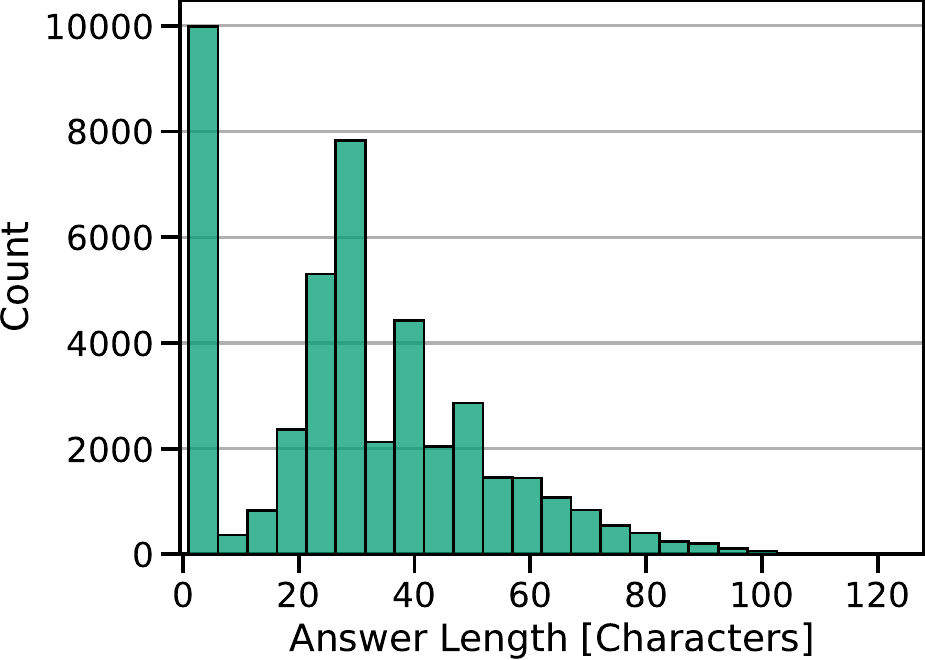}
    \caption{\textbf{\datasetName{}' question and open answer lengths are diverse.}}
    \vspace{-\baselineskip}
    \label{fig:question_length_dist}
\end{wrapfigure}
We took great care to ensure high diversity of the time series context and textual parts though sampling and diverse templating, while creating a sufficiently large dataset.
The samples are independent and identically distributed~(i.i.d.), which is a common assumption for data needed in machine learning.
We generated $30.000$ time series contexts of four consecutive actions, each, totaling $\sim$133.3~hours.
We pose five QA pairs for each of those contexts.
For best comparability between approaches adopting \datasetName{}, we split the dataset into predefined training, validation, and testing splits of $120,000$~($80\%$), $15,000$~($10\%$), and $15,000$~($10\%$) samples, respectively.
To prevent any leakage across splits, we ensure all five QA pairs per context time series are sorted into the same split.

For each unique sample, in addition to the question, answer, and motion trajectory, we also provide metadata, such as the question and answer types.
This equips \datasetName{} for diagnostics of TSQA models, e.g., aiding in understanding which settings prove the most difficult.
An optional fully textual version of the answers, as opposed to a binary Yes/No or multiple-choice A/B/C response, allows casting all tasks in an open format and provides additional training material.
For debugging and verifying models, we provide the ground-truth action sequence that was used to condition the time series generation.
Finally, we provide a textual description of the entire action sequence, again generated from diverse, manually reviewed templates.

When synthesizing the dataset, we ensure high diversity to obtain a challenging, varied dataset for training and evaluating systems.
\Cref{fig:question_length_dist}~(to the right) shows the wide range of question and open answer lengths.
Note that the high number of short answers stems from the counting tasks, where a good response is just the correct action count.
\Cref{fig:question-type-distribution} in \Cref{sec:app:details-generate-dataset} visualizes the final balanced distribution of both the question and answer types.
Similar to the textual components, diversity in the generated human motions is achieved through the STMC diffusion process, which produces organic, varied trajectories even for a fixed prompt.
\Cref{sec:app:details-generate-dataset} further investigates duplicates in \datasetName{} and shows that the presented generation procedure yields an effective sample size of >99\%.

\subsection{Metrics for Evaluating Answers}
\label{sec:dataset:metrics}

We grade the answers to the binary and multiple-choice questions using accuracy and the macro-averaged precision, recall, and F1 scores to account for any remaining class imbalances.
Answers that cannot be parsed as a valid answer to a binary or multiple-choice question, e.g., if not unambiguously recognizable as either yes or no, are counted as wrong.

Evaluation of the open questions necessitates comparing the textual reference answer to the answer given by the machine or human.
We thus employ the classic metrics ROUGE~\citep{linROUGEPackageAutomatic2004} and METEOR~\citep{banerjeeMETEORAutomaticMetric2005} commonly used to evaluate QA systems.
We omitted BERTScore~\citep{zhangBERTScoreEvaluatingText2020} because, since all questions are on similar topics, both good and bad responses yield uninformative high scores around $0.96$.
All these metrics, however, capture mostly syntactic similarity~\citep{imranLLaSALargeMultimodal2024}.
We thus augment them with a semantic comparison using LLMs as a judge comparing two textual answers~\citep{zhengJudgingLLMasajudgeMTbench2023,liLLMsasJudgesComprehensiveSurvey2024,guSurveyLLMasaJudge2025}.
We chose the following scores from 1 to 3 to obtain a nuanced assessment anchored in a well-defined scale:
\begin{enumerate}[label=\arabic*:,topsep=0.3\baselineskip,itemsep=0pt]
    \item The answer is terrible: The answer is factually incorrect or corrupt in some way (e.g., empty or nonsensical).
    \item The answer is bad: Somewhat correctly answers the question like the reference answer, yet contains additional useless/ambiguous/wrong information or is incomplete.
    \item The answer is good: Correctly answers the question and is helpful in the context of the scene.
\end{enumerate}

For a more intuitive interpretation, we follow common quality metrics, such as accuracy or F1 scores, by normalizing the range to $[0, 1]$, where 0 represents the poorest and 1 the best.
We aggregate the scores across multiple samples using the arithmetic mean.

As is good practice, we evaluate the quality of the LLM Judge by assessing its alignment with human judgments on a representative subset of 200 samples~\citep{bavarescoLLMsInsteadHuman2025}.
\Cref{sec:app:llmjudge} provides the detailed validation methodology, corresponding results, the full prompt, and detailed configuration.
We finally chose Qwen3~\citep[8B-AWQ with Reasoning,][]{yangQwen3TechnicalReport2025} due to its sufficient inference speed at excellent alignment with human judgment, specifically with a Pearson correlation coefficient $0.912$.
As an intuitive scale, it deviated on average by 7.45\%, permitting its use as a reliable measure of open-ended answer quality.

\section{On Solving \datasetName{}}
\label{sec:bench}

To validate the dataset and provide stable baselines, we conduct extensive experiments following the protocol defined in the previous \Cref{sec:dataset:metrics}.
Recall that \datasetName{} pairs motion time series with textual questions and target answers in binary, multiple-choice, and open formats.
We begin in \Cref{sec:bench:verify} with dataset verification and ablation studies to rule out trivial shortcuts~\citep{steinmannNavigatingShortcutsSpurious2024} and confirm that each component of the input (i.e., the time series context and question text) is indeed necessary.
In \Cref{sec:bench:humans}, we measure human performance on the same questions using auxiliary video renderings similar to \Cref{fig:motivational_chat}, and also test how reliably humans recognize the underlying actions.
Finally, in \Cref{sec:bench:baselines}, we evaluate three baselines of increasing capabilities: A naive LLM that sees the serialized numerical time series values and the question, the adapted SOTA time series LLM ChatTS~\citep{xieChatTSAligningTime2025}, and a simple neuro-symbolic pipeline~(xQA) built on top of purposefully structured information.
xQA is designed to demonstrate that the dataset can be answered by TSQA models, but that current TS-LLMs still require explicit structure.

\subsection{Dataset Verification and Ablations}
\label{sec:bench:verify}

We first ensure the logical consistency of \datasetName{} by answering its questions based on ground-truth descriptions of the action sequences.
Second, we investigate whether the questions alone are sufficient to answer them, which should not be possible in a carefully constructed dataset.
Lastly, we attempt to determine whether LLMs can solve \datasetName{} using only the time series, without the questions, again ruling out any shortcuts.

We aim to answer the following set of questions:
\begin{enumerate}[label=\emph{(Q\arabic*)}:,topsep=0.3\baselineskip,itemsep=0pt,leftmargin=8ex]
    \item Does the action sequence provide sufficient information to answer the questions?
    \item Are there shortcuts between the questions and answers?
    \item Are there shortcuts between the time series and answers?
\end{enumerate}

We base our results on the widely used Llama~3.1~8B model~\citep{llama-teamLlama3Herd2024} \emph{finetuned} on the whole train split.
This model size offers a good balance of reasoning capabilities and the possibility to fine-tune it.
See \Cref{sec:app:eval-details:llms} for implementation details, and \Cref{sec:app:eval-details:lm_retention} for how this deteriorates general language modeling capabilities.

\paragraph{Ground-Truth Scene Descriptions and Questions~\emph{(Q1)}.}
In this experiment, we evaluate the model performance using textual ground-truth descriptions of the action sequence, which are only available for model debugging, together with the corresponding question as input.
This setup acts as a check for soundness, verifying that the dataset is logically consistent and that the questions are answerable given the provided context.

The accuracy and F1 score on the binary variant were $99.10\%$ and $99.12\%$, and for the multi variant $99.98\%$ and $99.98\%$, respectively.
Furthermore, the very good results for the open-ended questions serve as an aspirational upper bound for subsequent experiments, with an LLMJudge score of $0.9434$.
The ROUGE and METEOR scores were $0.3380$ and $0.5099$, respectively, indicating a decent syntactic alignment with the ground truth.
The excellent performance for all three variants confirms the well-formedness of \datasetName{} and affirms \emph{(Q1)}.
Moreover, this validates that Llama~3.1~8B has sufficient capacity for these experiments.

\paragraph{Only the Questions~\emph{(Q2)}.}
Next, we verify that the questions cannot be answered independently of the time series context, which we expect to be impossible if no shortcuts are present.
In other words, the accuracy should be reasonably close to random guessing.

The results for the \emph{Only Question} ablation in \Cref{tab:main-results} show that responses across all three answer types are close to random guessing.
We thus reject \emph{(Q2)} as we did not find any spurious correlations.

\paragraph{Only the Time Series Encoded as Text~\emph{(Q3)}.}
We assess how much useful information can be extracted from the motion trajectory when the textual question is withheld.
The trajectory is serialized as text following \citet{gruverLargeLanguageModels2023}; preprocessing and prompt details are provided in \Cref{sec:app:eval-details:llms}.

The time-series-only model reaches a binary accuracy of $49.21\%$ and multiple-choice accuracy of $33.29\%$, both close to uniform guessing, with macro F1 scores of $33.62\%$ and $20.39\%$, respectively~(\Cref{tab:main-results}).
No closed-answer prediction is invalid under the parser used for these configurations.
Open-answer performance is low, with an LLMJudge score of $0.1082$.
Because several different questions share the same trajectory, the missing question makes the requested answer ambiguous.
In summary, we can rule out any substantial confounders between the time series and answers in \datasetName{} \emph{(Q3)}.

We proceed by evaluating human responses to the questions before baseline models.

\begin{table}[p]
    \centering
    \caption{\textbf{Dataset variants and ablations show that \datasetName{} is well-formed. Furthermore, humans can successfully solve the binary and multi splits. The Naive Baseline~(except for open) and ChatTS struggle to solve most splits of the \datasetName{} dataset, whereas xQA successfully leverages the additional information.}
    Human, ChatTS, and xQA rows provide reference scores under their respective evaluation setups; direct comparison is only possible within the sections marked by the gray dividers~(\textcolor{gray}{\textbf{---}}).
    $n$ indicates the total number of human responses.
    GT denotes ground-truth action labels and AE a supervised action encoder using pre-cut segments.
    More details on humans and xQA are provided in \Cref{sec:app:results:details}.}
    \label{tab:main-results}
    \denseColumns
    \begin{tabular}{clcccc}
        \toprule
         & System                    & Accuracy (\textuparrow) & Precision (\textuparrow) & Recall (\textuparrow) & F1 (\textuparrow) \\
        \midrule
        \multirow{9}{*}{\textbf{Binary}}
         & Ablation: Only Question & 64.53\% & 65.32\% & 64.70\% & 64.22\% \\
         & Ablation: Only TS & 49.21\% & 50.15\% & 50.00\% & 33.62\% \\
         \subtleCmidrule{\cmidrule{2-6}}
         & Humans ($n=840$)          & 77.50\%                 & 81.08\%                  & 74.66\%               & 77.74\% \\
         \subtleCmidrule{\cmidrule{2-6}}
         & Naive: TS + Question & 62.94\% & 64.54\% & 63.19\% & 62.16\% \\
         & ChatTS     & 51.69\% & 55.38\% & 25.36\% & 34.79\% \\
         \subtleCmidrule{\cmidrule{2-6}}
         & xQA-Llama on GT & 79.44\% & 84.91\% & 72.42\% & 78.17\% \\
         & xQA-Qwen on GT     & 80.92\% & 92.28\% & 68.14\% & 78.39\% \\
         & xQA-Llama on AE & 79.46\% & 84.60\% & 72.84\% & 78.28\% \\
         & xQA-Qwen on AE     & 81.00\% & 92.52\% & 68.10\% & 78.46\% \\
        \midrule
        \multirow{9}{*}{\textbf{Multi}}
         & Ablation: Only Question & 33.32\% & 30.38\% & 33.00\% & 26.25\% \\
         & Ablation: Only TS & 33.29\% & 21.94\% & 33.08\% & 20.39\% \\
         \subtleCmidrule{\cmidrule{2-6}}
         & Humans ($n=820$)          & 86.59\%                 & 86.55\%                  & 86.61\%               & 86.54\% \\
         \subtleCmidrule{\cmidrule{2-6}}
         & Naive: TS + Question & 34.01\% & 22.74\% & 33.68\% & 26.62\% \\
         & ChatTS     & 30.40\% & 30.13\% & 30.58\% & 29.12\% \\
         \subtleCmidrule{\cmidrule{2-6}}
         & xQA-Llama on GT & 81.50\% & 81.64\% & 81.51\% & 81.50\% \\
         & xQA-Qwen on GT     & 88.01\% & 88.04\% & 88.03\% & 88.01\% \\
         & xQA-Llama on AE & 81.18\% & 81.30\% & 81.19\% & 81.18\% \\
         & xQA-Qwen on AE     & 87.97\% & 87.99\% & 87.98\% & 87.97\% \\
        \midrule[\heavyrulewidth]
        \rule{0pt}{3ex}
         & System & ROUGE (\textuparrow) & METEOR (\textuparrow) & \multicolumn{2}{c}{LLMJudge (\textuparrow)} \\
        \midrule
        \multirow{9}{*}{\textbf{Open}}
         & Ablation: Only Question & 0.4233 & 0.2838 & \multicolumn{2}{c}{0.2092} \\
         & Ablation: Only TS & 0.0956 & 0.0491 & \multicolumn{2}{c}{0.1082} \\
         \subtleCmidrule{\cmidrule{2-6}}
         & Humans ($n=440$)        & 0.0953 & 0.2635 & \multicolumn{2}{c}{0.6148} \\
         \subtleCmidrule{\cmidrule{2-6}}
         & Naive: TS + Question & 0.8199 & 0.6947 & \multicolumn{2}{c}{0.8759} \\
         & ChatTS                  & 0.0455 & 0.1060 & \multicolumn{2}{c}{0.0702} \\
         \subtleCmidrule{\cmidrule{2-6}}
         & xQA-Llama on GT         & 0.0913 & 0.3677 & \multicolumn{2}{c}{0.7729} \\
         & xQA-Qwen on GT          & 0.1789 & 0.3697 & \multicolumn{2}{c}{0.8279} \\
         & xQA-Llama on AE         & 0.0898 & 0.3679 & \multicolumn{2}{c}{0.7753} \\
         & xQA-Qwen on AE          & 0.1782 & 0.3692 & \multicolumn{2}{c}{0.8176} \\
        \bottomrule
    \end{tabular}
\end{table}

\subsection{Human Performance as Reference}
\label{sec:bench:humans}

In the preceding \Cref{sec:bench:verify}, we first demonstrated that \datasetName{} contains sufficient information to be solved, provided that actions can be identified correctly, and second, ensured that handling both TS and text modalities is indeed required.
Next, we will investigate whether humans can solve the tasks, given that most can rather reliably identify actions in day-to-day life.
Furthermore, this permits judging model performance relative to humans, thereby providing an intuitive scale of progress for the emerging field of TSQA.

\paragraph{Setup.}
We performed a separate study for each of the three splits~(binary, multi, and open), with matching answer formats and the respective evaluation metrics from \Cref{sec:dataset:metrics}.
We hired a diverse cohort of participants through \emph{Prolific Academic}\footnote{\url{https://www.prolific.com}}, an established crowdsourcing platform for behavioral research.
Participants were laypeople, but selected to be native speakers of English to rule out language barriers as potential sources of error.
In the open variant, we additionally verify if humans can correctly identify the individual actions.
See \Cref{sec:app:eval-details:human} for details on the study and demographics of its participants.

\paragraph{Results.}
The results in \Cref{tab:main-results} generally confirm that humans can solve the dataset in the provided format, especially for the binary and multi splits.
Specifically, the F1 scores---while not quite reaching the upper ceiling from \Cref{sec:bench:verify}~(Q1)---are overall good.
The scores for the binary variant are slightly lower than those for the multiple-choice variant, possibly because participants ponder less when fewer options and guidance are presented.
For the open variant, we measured a moderate LLMJudge score of $0.6148$.
We suspect that the lower performance compared to other answer types partly stems from online annotators being less ambitious in their textual answers due to the increased effort relative to fixed answer types.

Overall, the gap between the results for LLMs trained on ground-truth scene descriptions from \Cref{sec:bench:verify} and these human results likely stems from the added task of identifying actions.
For instance, \texttt{catching a ball} is commonly misidentified as \texttt{throwing a ball}.
The details provided in \Cref{sec:app:eval-details:human}, however, overall confirm that humans can effortlessly discern most actions and solve \datasetName{}.
Further analysis is provided in \Cref{sec:app:results:human_detection_and_perf}.

\subsection{Baselines}
\label{sec:bench:baselines}

To illustrate what our dataset reveals about current TSQA methods, we evaluate three complementary baselines:
\begin{enumerate}[label=(\arabic*), nosep, labelindent=1em, leftmargin=*, topsep=0pt, partopsep=0pt, parsep=0pt, itemsep=0pt]
  \item A general-purpose fine-tuned text model receives the serialized numerical time series alongside the question, without a dedicated time-series encoder.
  \item ChatTS~\citep{xieChatTSAligningTime2025} represents purpose-built TS-LLMs and is designed for interactions contextualized by multivariate numerical data.
  \item xQA is a neuro-symbolic pipeline that augments a general-purpose LLM with structured action information extracted from the time series. Its variants help separate action recognition from question answering.
\end{enumerate}
Together, these systems characterize different approaches to solving \datasetName{}' TSQA tasks.

\paragraph{(1)~Naive Baseline: Time Series Encoded as Text and Questions.}
Following the setup of the verification experiments in \Cref{sec:bench:verify}, we finetune Llama~3.1~8B on \datasetName{}.
The time series is serialized to text as in the \enquote{Only TS} ablation and then concatenated with the question and an introductory instruction.
The full prompt is shown in \Cref{sec:app:eval-details:llms}.

\paragraph{(2)~Adapting ChatTS.}
We include ChatTS because it is specifically designed to tokenize and condition on time-series inputs, making it an important state-of-the-art baseline for our task.
ChatTS first encodes time series into tokens with attached metadata, such as the number of time steps and normalization coefficients.
It then injects these into the prompt, which is fed to a specially finetuned language model to perform standard auto-regressive next-token prediction.

However, the released model is limited to 30 variates and therefore cannot directly process the 72 ones in \datasetName{}.
To still facilitate a comparison, we only pass 10 particularly distinctive joints~(with three coordinates each) to ChatTS, as detailed in \Cref{sec:app:eval-details:chatts}.
This restriction of the data channels is simultaneously a potentially advantageous human prior.

\paragraph{(3)~xLSTM-based QA Systems: xQA-Qwen \& xQA-Llama.}
End-to-end multimodal systems for TSQA typically require finetuning on substantial amounts of data.
Instead, we separate perception~(neural) from reasoning~(symbolic) by passing only a short sequence of textual task descriptions and discrete action labels to an LLM without retraining.
To facilitate this, we train a segmentation model on ground-truth action labels of the action time sequence, which are typically \emph{not} accessible to models solving the dataset.

Let the input to \emph{xQA} be a multivariate time series $\bm{x}_{1:T}\in\R^{T\times V}$ partitioned into $M=4$ equal segments for each action label, with boundaries $(s_i,e_i)$ for $i \in \{1,\dots,M\}$.
We denote the finite action vocabulary by $\mathcal{A}$ (with $|\mathcal{A}|=19$, see \Cref{sec:app:details-generate-dataset}).
We use xLSTM-Mixer~\citep{kraus2025xlstmmixer} as an efficient time-series embedding backbone, followed by a lightweight supervised classification head predicting the action label per segment.
Specifically, for $i$, the model predicts $a_i = f_\theta\left(\bm{x}_{s_i:e_i}\right)\in\mathcal{A}$, yielding an action sequence $\bm{a}=(a_1,\dots,a_M)$.

Reasoning is then delegated to an instruction-tuned LLM $g$ consuming an instruction $I$, the action sequence $\bm a$, the question $q$, and two few-shot demonstrations $D_1$ and $D_2$, yielding a response $r = g\left( I, \bm{a}, q; D_1, D_2 \right)$.
The instruction $I$ requires a concise, numbered reasoning chain followed by a final answer generated through constrained generation.

To disentangle the evaluation of perception from reasoning, we report the performance of xQA on both the action encoder~(AE) and the ground-truth~(GT) action labels.
xQA-AE serves as a supervised diagnostic baseline jointly performing both perception and reasoning, while xQA-GT provides an oracle-style reasoning ceiling for a given LLM.
Both leverage privileged action-sequence information not available to standard methods for solving QuAnTS.

Because the interface is text-only, we instantiate $g$ with the LLMs Llama~3.1~8B and Qwen3 8B \emph{without any finetuning}.
Implementation details are provided in \Cref{sec:app:eval-details:nesy}.

\paragraph{Results.}
The Naive Baseline shows substantially different behavior across answer formats~(\Cref{tab:main-results}).
For binary answers, it achieves an F1 score of $62.16\%$, which is below the performance of the question-only ablation of $64.22\%$.
For multiple-choice answers, accuracy is close to uniform guessing with an F1 of $26.62\%$.
All binary and multiple-choice answers were valid when parsing.
For open answers, however, the Naive Baseline attains an LLMJudge score of $0.8759$, indicating that a QLoRA-adapted text model can produce accurate open answers from serialized trajectories and questions in this setting.

However, this does not demonstrate general time series understanding, as this lightweight model adaptation causes other model capabilities to collapse entirely.
Specifically, in \Cref{sec:app:eval-details:lm_retention}, we measure the change in general language modeling abilities resulting from low-rank model adaptation.
We find that perplexity deteriorates drastically: on WikiText-2~\citep{merity2016pointer}, it increases by seven orders of magnitude, where lower would be better~(\Cref{tab:language-model-deterioration}).
This mirrors the findings for the ablations \enquote{Only Question} and \enquote{Only TS} from \Cref{sec:bench:verify}~(see \Cref{sec:app:eval-details:lm_retention} for details).
In contrast, all other baselines retain their language modeling capabilities.

ChatTS does not reliably answer the questions of \datasetName{} in the evaluated setup.
Its accuracy is close to uniform guessing for the binary and multiple-choice variants, and its open-answer LLMJudge score is $0.0702$.
Note that ChatTS had access to only 10 of 24 joints due to model limitations~(see \Cref{sec:app:eval-details:chatts}), which both limited the information available and assisted through an expert-informed focus.

xQA achieves strong scores using explicit action-sequence information, with Qwen3~8B consistently ahead of Llama~3.1~8B in the reported results.
The similar performance with action-encoder and ground-truth labels indicates that the supervised action encoder can provide useful inputs to the downstream question-answering model~(see also \Cref{fig:action_encoder_performance} in \Cref{sec:app:eval-details:nesy}).

Recall that xQA uses privileged action supervision unavailable to the Naive and input-ablation models.
Its results, therefore, characterize the potential of hybrid models rather than being directly comparable to the other baselines.
A comparison to human performance must be made cautiously because of a mismatch in input modalities due to the videos shown to the human participants.

\section{Conclusion and Outlook}
\label{sec:conclusion}

There is very limited research on handling complex combinations of modalities, such as time series question answering.
To foster research into models with new capabilities, we propose \datasetName{}, a novel dataset featuring diverse questions and answers on human motion trajectories in binary, multiple-choice, and open-ended variants.
We provide open access to the flexible generation pipeline and dataset and verify its well-formedness and challenges through extensive experiments.
Additionally, the baseline performance of three different models provides key insights into the models' capabilities and limitations, highlighting the need for dedicated models.
Lastly, we contextualize the state of learned systems by reporting human performance on \datasetName{}.

\subsection{Limitations}

The main limitation of this work stems from its focus on a single, controlled domain: QA on human motion data.
While this provides an easily debuggable diagnostic setting, it does not yet provide a benchmark for generic TSQA.
By extension, the xQA baseline serves as a diagnostic tool for \datasetName{}, but might be challenging to transfer to arbitrary domains.

The time series features do not capture the full range of all human motion and are specific to the signatures generated by STMC.
Generally, the structure is also restricted to 19 actions, 4 fixed-length slots each, a fixed set of question/answer templates, a fixed overall sequence duration, regular sampling, and a lack of real-world sensor issues.

We employed LLM judges to assess and define the quality metrics for \datasetName{}.
While it was evaluated thoroughly in \Cref{sec:dataset:metrics,sec:app:llmjudge}, some biases may persist, especially given that some baselines and the judge share the same foundation model.

As identified by \Cref{sec:bench:humans,sec:app:eval-details:human}, \datasetName{} is a challenging task even for humans, with them falling far short of the respective perfect scores.
Extra care is thus needed when interpreting results.
Reaching human levels in both language modeling and simultaneous time series understanding would already constitute a substantial step forward.

\subsection{Future Work}

Moving forward, we identify two primary avenues for building on this work.
First, regarding the dataset itself, the presented scope of TSQA could be broadened to include time series from other application domains and more complex, compositional question types.
In practice, one will likely combine data about generic time series knowledge (such as identifying peaks and seasonal patterns) with domain-specific data (such as the specifics of human motion).
Investigating how to curate optimal mixtures of them remains an open question.

This leads to the second key path forward: building truly multimodal models for time series and text.
Our results highlight the urgent need for improved architectures specifically designed to address the challenges of TSQA.
Firstly, substantially more work is needed to equip pre-trained models with the capabilities to reason about time series in zero-shot settings.
Furthermore, moving from single-turn QA to longer conversational systems would open up many new fields of application.
It remains unclear to what extent this requires dedicated multi-turn multimodal datasets, or whether conversational abilities can be fully retained in post-trained models.

\pagebreak

\FloatBarrier

\impact{%
Time series data occurs in many real-world applications---often in tight relation to our daily lives.
Datasets like \datasetName{} work towards models that provide better access to this data modality through more intuitive textual interfaces.
This enables many applications with a positive impact, including personalized health recommendations during exercise, insights into sleep quality, personal finance recommendations, and doctors more effectively summarizing complex measurements from patients.
On the flip side, some of these fields touch upon very intimate parts of our lives, necessitating a thorough assessment of their potential invasions of privacy.
As with any machine learning system deployed to individuals, great care must be taken to adequately represent minorities to ensure a fair model and to clarify the capabilities and limitations of such systems.
If applied to less personalized data, such as industrial manufacturing or computer networking logs, we anticipate the potential gains in accessibility and integration to far outweigh the limited potential harm.

Thanks to the entirely synthetic dataset, we do not foresee any major adverse consequences regarding \datasetName{} itself.
While the STMC human motion generation was trained on data obtained from human subjects, these respective datasets are anonymized and widely used, and, to the best of our knowledge, no major concerns have surfaced so far.
Since it might not represent people with different cultural backgrounds, certain (motoric) limitations, or body morphologies, real-world applications should carefully assess their representation of the intended target audiences.

Finally, we stress that \datasetName{} is intended mainly as a data source and benchmark of TSQA capabilities, not as a deployment-ready system.
We attach a comprehensive datasheet as \Cref{sec:app:datasheet}.}

\acks{%
We cordially thank Clara Moos and Patrick Goldbeck for their help in drafting the initial iterations of the text generation process.

This work received funding from the German Federal Ministry of Research, Technology and Space~(BMFTR) project \enquote{KompAKI} within the \enquote{The Future of Value Creation -- Research on Production, Services and Work} program~(funding number 02L19C150), managed by the Project Management Agency Karlsruhe~(PTKA), the ACATIS Investment KVG mbH project \enquote{Temporal Machine Learning for Long-Term Value Investing}, and the EU project EXPLAIN, under the BMFTR~(grant 01-S22030D).
The project also benefited from the early stage of the German Federal Ministry for Economic Affairs and Energy project \enquote{Souveräne KI für Europa}~(13IPC040G) as part of the EU funding program IPCEI-CIS.
Devendra Singh Dhami received support from the Department of Mathematics and Computer Science at Eindhoven University of Technology and the Eindhoven Artificial Intelligence Systems Institute.
Flora Salim would like to acknowledge the support of Cisco’s National Industry Innovation Network~(NIIN) Research Chair Program.
The authors are responsible for the content of this publication.
}

\clearpage
\bibliography{references}

\clearpage
\appendix
\section{Details on Dataset Generation}
\label{sec:app:details-generate-dataset}

\paragraph{Action Sequences.}
To ensure actions are repeated sufficiently often to permit asking, for instance, interesting counting questions, we double the chance of duplicating actions in each activity sequence.
We chose the following 19 clearly discernible actions to generate the prompts:
\texttt{playing guitar},
\texttt{punching},
\texttt{running},
\texttt{jumping once},
\texttt{drinking with the left hand},
\texttt{bowing},
\texttt{throwing a ball},
\texttt{shaking hands},
\texttt{skipping rope},
\texttt{picking something up with both hands},
\texttt{dancing},
\texttt{eating with the right hand},
\texttt{waving},
\texttt{catching a ball}, \\
\texttt{kicking a ball},
\texttt{sitting down},
\texttt{golfing~(swinging a club)},
\texttt{T-posing}, and
\texttt{holding a baby}.
We targeted an equal distribution of question and answer types.
Furthermore, we ensured that the correct answer~(in case of binary and multi) sits at a randomized position.

\paragraph{Configuration of Human Motion Generation.}
We experimented with several different STMC configurations to discover the most natural movement.
We found that prompting all body parts jointly with a guidance scale of $3.0$ and an overlap of $2l = 1.0$ seconds was most effective.
As quantitatively evaluated later, this produced a diverse set of unique time series.

\paragraph{Alternative Human Motion Generation Systems.}
We also experimented with \emph{Motion Diffusion Model}~(MDM) \citep{tevetHumanMotionDiffusion2023} using autoregressive stitching by conditioning on previously generated subsequences, and investigated using its successor \emph{PriorMDM}~\citep{shafirHumanMotionDiffusion2024}.
However, neither reached the perceptual quality of STMC.

\begin{figure}[p]
    \centering
    \includegraphics[width=\linewidth]{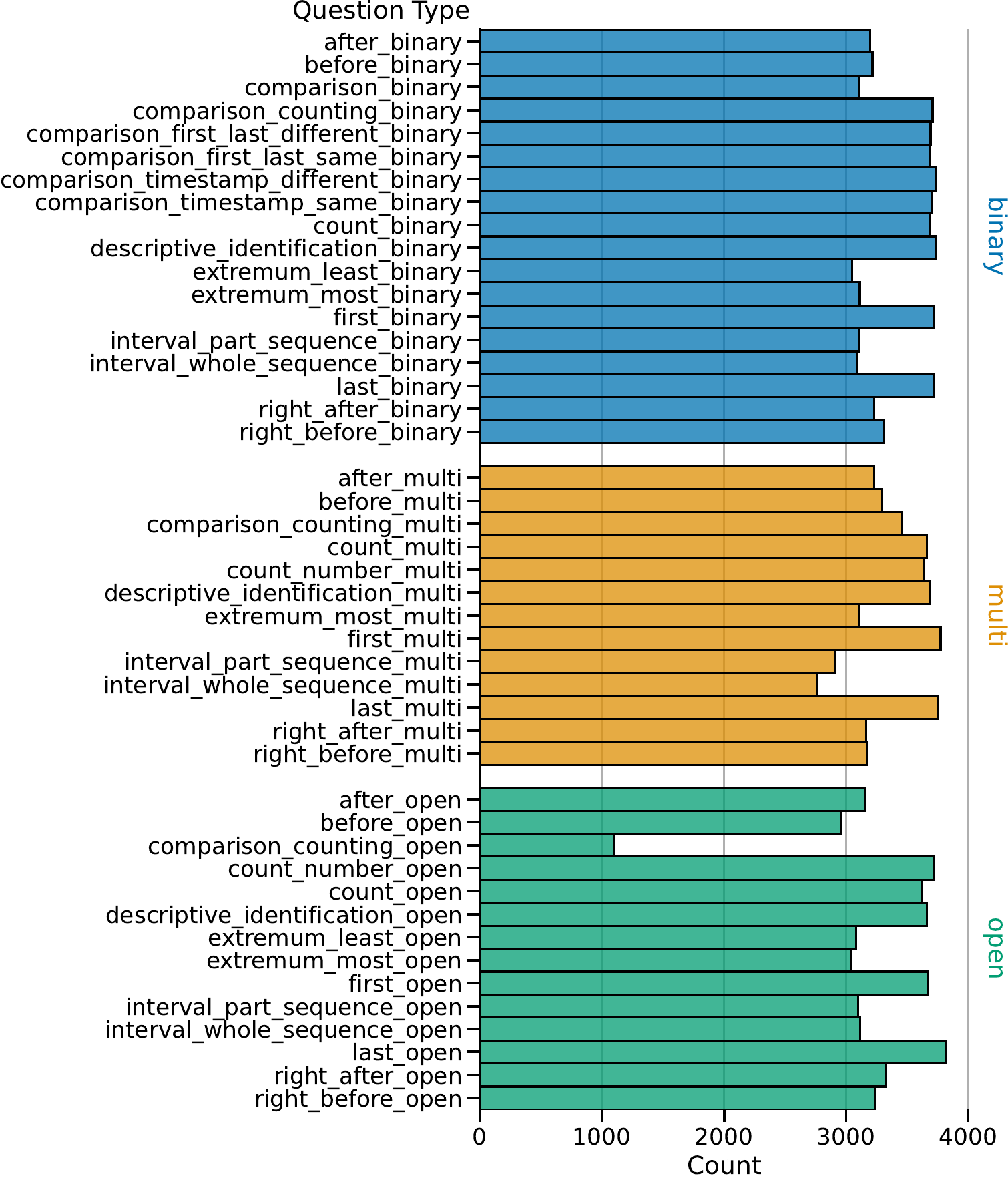}
    \caption{\textbf{\datasetName{} features a diverse and balanced question and answer types distribution.}
    Some categories, such as comparison\_counting\_open, occur less often because of the particular circumstances that allow them to be posed without trivial responses.}
    \label{fig:question-type-distribution}
\end{figure}

\paragraph{Diverseness.}
The question and answer types distribution is visualized in \Cref{fig:question-type-distribution}.
To ensure that almost all instances in the dataset are unique, we also searched for any duplicates.
\Cref{tab:uniqueness_text} shows that even if there exist multiple instances with the same textual question and answer, they refer to different action labels.
Moreover, there is not a single time series that is exactly duplicated in the dataset (checked with SHA-256).
However, we also want to avoid overly similar motions, not just exact matches.

Each QuAnTS sample carries a context trajectory $x \in \R^{T \times V}$ with $T = 320$ time steps and $V = 72$ variates. Each context is shared across five QA pairs, leaving $n = 30{,}000$ distinct series.
We first define the scale at which to compare the variate $v$ of two time series $x, y$ as 
$$
    s_v = \max\left( 10^{-8},
        \sqrt{\tfrac{1}{2}\left(\Var_t[x_v] + \Var_t[y_v]\right)}
    \right),
$$
using the unbiased variance over the time axis.
Pooling $s_v$ per channel makes the score invariant to scales of each joint axis.
The distance of $x$ and $y$ is then defined as
$$
    d(x,y) = \max_{v \in 1,\dots,V}  \operatorname{Q}_{0.95}^t
        \left( \frac{\lvert x_{t,v} - y_{t,v} \rvert}{s_v} \right) .
$$
This distance is intuitively thought of in units of channel standard deviations $\sigma$: $d(x,y) > \tau$ means that in at least one channel, at least 5\% of time steps differ by more than $\tau$ pooled standard deviations.
Because the outer reduction is a maximum over channels, two time series that differ on a subset of channels differ overall.
We therefore first rule out duplicates on a subset of 8 channels, then on the complete set.

\begin{figure}[ht]
    \centering
    \includegraphics[width=\linewidth]{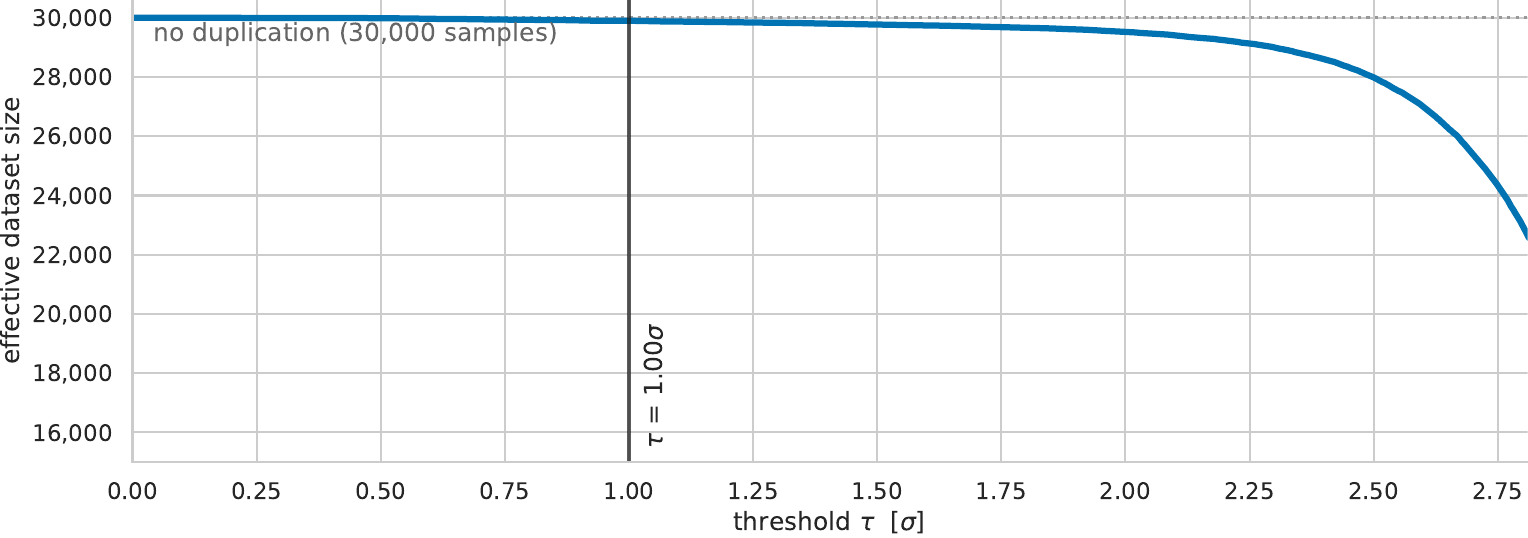}
    \caption{\textbf{Effectively all generated numerical motion time series are unique.}
    At $\tau = 1 \cdot \sigma$, the effective dataset size under soft equality is 29,895 (99.65\% of the full 30,000).}
    \label{fig:uniqueness_time_series}
\end{figure}

We finally compute groups of redundant time series using conservative single-linkage clustering, i.e., by transitively grouping all pairs of similar series.
Retaining only a single member per such cluster again yields an effective sample size.
\Cref{fig:uniqueness_time_series} shows that under the defined soft equality, almost no duplicates exist.

\begin{table}[ht]
    \centering
    \caption{\textbf{Effectively all combinations of question-answer pairs and action sequence labels are distinct.}
    Furthermore, the probabilistic motion generation did not produce a single time series duplicate.
    All checks were performed using exact matching.}
    \label{tab:uniqueness_text}
    \begin{tabular}{llccc}
        \toprule
        Data & Answer Type & \# Samples & Eff. \# Samples & Eff. Samples \\
        \midrule
        \multirow[c]{3}{*}{\shortstack[l]{Question +\\Answer}}
            & Binary & 61,804 & 45,253 & 73.220\% \\
            & Multi  & 43,599 & 41,021 & 94.087\% \\
            & Open   & 44,597 & 33,194 & 74.431\% \\
        \midrule
        \multirow[c]{3}{*}{\shortstack[l]{Question +\\Answer +\\Action Labels}}
            & Binary & 61,804 & 61,791 & 99.979\% \\
            & Multi  & 43,599 & 43,598 & 99.998\% \\
            & Open   & 44,597 & 44,540 & 99.872\% \\
        \midrule
        Time Series & --- & 30,000 & 30,000 & 100.00\% \\
        \bottomrule
    \end{tabular}
\end{table}

\section{Evaluating the LLM Judge}
\label{sec:app:llmjudge}

\Cref{sec:dataset:metrics} described the metrics used to evaluate the answers of the machine learning models and humans.
This section contains the full details for evaluating the LLM Judge.

\paragraph{Choice and Configuration of the LLM.}
We evaluated several moderately sized language models to limit the energy consumption and inference time required for the evaluation, while also considering that this evaluation scheme might be replicated by other works adopting \datasetName{}.
Namely, we tried three models of the Qwen3 family (8B-AWQ with Reasoning, 8B-AWQ without reasoning, and 4B-AWQ without reasoning) \citep{yangQwen3TechnicalReport2025} and Llama 3.2 (3B, instruction tuned) \citep{llama-teamLlama3Herd2024}, where AWQ stands for Activation-aware Weight Quantization~\citep{linAWQActivationawareWeight2025}.
The sampling parameters used to generate judge scores from the four LLMs are shown in \Cref{tab:sampling-params-llmjudge}.
To allocate sufficient space to intermediate \enquote{thinking} artifacts, we generate up to $8192$ tokens.

\begin{table}[ht]
    \centering
    \caption{\textbf{Sampling parameters used across the four LLM judge candidates.}}
    \label{tab:sampling-params-llmjudge}
    \begin{tabular}{lcccc}
        \toprule
        Model                   & Temperature & Top-$p$ & Top-$k$ & Repetition Penalty \\
        \midrule
        Qwen3 4B/8B (Reasoning) & 0.6         & 0.95    & 20      & 1.5                \\
        Qwen3 4B                & 0.7         & 0.8     & 20      & 1.5                \\
        Llama 3.2 3B (Instruct) & 0.8         & 0.8     & 20      & 1.5                \\
        \bottomrule
    \end{tabular}
\end{table}

\pagebreak

\paragraph{Prompt and Constrained Generation.}
To ensure a reliably parsable output, we explain the format in the prompt and constrain generated tokens to adhere strictly to the described JSON format using SGLang~\citep{zhengSGLangEfficientExecution2024}.
The rationale was allowed to be up to 500 characters long.
We used the following prompt:

\begin{llmprompt}{User Prompt for the LLM Judge}
    You will be given a scene\_description with timestamps and a scene\_question regarding that scene. You will also be given a reference\_answer and system\_answer couple.
    Your task is to provide a TotalRating scoring how well the system\_answer answers the user concerns expressed in the scene\_question. The reference\_answer is provided as reference for a very good answer.
    Give your answer on a scale of 1 to 3, where 1 means that the system\_answer is not helpful at all, and 3 means that the system\_answer completely and helpfully addresses the scene\_question.
    You will also provide a brief rationale for your rating.
    \bigskip

    Here is the scale you should use to build your answer:\\
    1: The system\_answer is terrible: the answer is factually incorrect or corrupt in some way (e.g., empty or nonsensical)\\
    2: The system\_answer is bad: somewhat correctly answers the scene\_question like the reference\_answer, yet contains additional useless/ambiguous/wrong information or is incomplete\\
    3: The system\_answer is good: correctly answers the scene\_question and is helpful in the context of the scene
    \bigskip

    Provide your assessment in the following JSON format:
    \begin{lstlisting}
{
    "BriefRationale": "<your rationale, at most 3 sentences>",
    "TotalRating": <your rating>
}
\end{lstlisting}

    You MUST provide values for "BriefRationale" and "TotalRating" in your answer.
    Do NOT respond with any other text in your answer.
    Immediately, start with the JSON object, without any additional text or explanation.
    Make sure to output complete and valid JSON.
    \bigskip

    Now here are the question and answer:\\
    - scene\_description: \{scene\_description\}\\
    - scene\_question: \{scene\_question\}\\
    - reference\_answer: \{reference\_answer\}\\
    - system\_answer: \{system\_answer\}
    \bigskip

    Provide your assessment.
\end{llmprompt}

\pagebreak

\paragraph{Verifying Alignment with Human Judgment.}
To ensure that the LLM Judge aligns well with human assessment, we collected 200 random output samples across the experiments from \Cref{sec:bench:verify}.
Three human annotators were presented with the question, a reference answer, and the model answer.
Their agreement with each other was very high, with 88.5\% perfect alignment at an average variance of $\sim$0.04 across annotators.
We can thus employ the mean of the human ordinal scale as a baseline to evaluate the LLM Judges against.
All four language models are evaluated in \Cref{tab:llm-judge-results}.
Due to the high quality of responses at reasonable inference speed, we used Qwen3~8B~(AWQ, Reasoning) for the LLM Judge.

\begin{table}[ht]
    \centering
    \caption{\textbf{Qwen3~8B~(AWQ, Reasoning) aligns best with human judgement.} 
        Highlighted \textbf{bold} is the best value per column, with the arrows (\textdownarrow/\textuparrow) indicating which direction is desired.
        Average over 10 seeds.}
    \label{tab:llm-judge-results}
    \small
    \begin{tabular}{lcccc}
        \toprule
        \textbf{Model}            & \textbf{Avg. $\Delta$ (\textdownarrow)} & \textbf{Pearson (\textuparrow)} & \textbf{Spearman (\textuparrow)} & \textbf{Kendall (\textuparrow)} \\
        \midrule
        Qwen3 8B (AWQ, Reasoning) & \textbf{0.149}                          & \textbf{0.912}                  & \textbf{0.927}                   & \textbf{0.878}                  \\
        Qwen3 8B (AWQ)            & 0.254                                   & 0.843                           & 0.857                            & 0.794                           \\
        Qwen3 4B (AWQ)            & 0.155                                   & 0.883                           & 0.891                            & 0.853                           \\
        Llama 3.2 3B (Instruct)   & 0.430                                   & 0.640                           & 0.654                            & 0.603                           \\
        \bottomrule
    \end{tabular}
\end{table}

\section{Details on Evaluating \datasetName{}}
\label{sec:app:eval-details}

\subsection{Verifying and Benchmarking using Language Models}
\label{sec:app:eval-details:llms}

In \Cref{sec:bench}, we finetune Llama~3.1~8B independently for the question-only, time series-only, and Naive configurations and each answer format, yielding nine QLoRA adapters.
Each adapter uses its corresponding training split for five epochs, with an effective batch size of 16 and a fixed seed of 42.
We use AdamW~\citep{loshchilovDecoupledWeightDecay2018} with $\beta=(0.9, 0.999)$, weight decay $0.01$, learning rate $2\times10^{-4}$ with linear decay, no warmup, and gradient clipping at $1.0$.
QLoRA~\citep{dettmers2023qlora} uses rank 16, $\alpha=32$, dropout $0.1$, and no trainable biases.
Adapters target the attention and MLP projections; base weights remain frozen.
We use NF4 double quantization with BF16 computation and gradient checkpointing.

The time series undergo per-trajectory min-max scaling over the complete tensor to $[-999,+999]$, rounding to integers, and serialization as compact nested JSON preserving the $320 \times 24 \times 3$ rank-3 structure.

Training supervises only answer tokens and the final end-of-sequence token, without a chat template or sequence packing.
The final fifth-epoch adapters generate greedily without output constraints, allowing up to 32 new tokens for closed answers and 128 for open answers.
Invalid closed answers count as incorrect; the main results use the held-out test split.

Finetuning and evaluating the nine adapters, including prediction generation, open-answer judging, and perplexity evaluation, required approximately $5{,}492$ Nvidia H100 GPU-hours due to the very long context length.

\pagebreak
\subsubsection{Prompts for \Cref{sec:bench:verify}.}
\vspace{-0.5\baselineskip}

\newcommand{\trainingTarget}[1]{\textcolor{orange}{#1}}

The following prompts were used to answer the questions in \Cref{sec:bench}.
\trainingTarget{Orange} denotes the target answer, which is supplied only during training and omitted from the inference prompt.
For the question-only, time series-only, and Naive adapters, the answer and end-of-sequence tokens are supervised, conditioned on the preceding prompt.

\vspace{-0.5\baselineskip}

\begin{llmprompt}{Prompt For Ground Truth Scene Descriptions and Questions (Q1)}
    \#\#\# Context: \{textual description of the action sequence\}
    \bigskip

    \#\#\# Question: \{question\}
    \bigskip

    \#\#\# Answer: \trainingTarget{\{answer\}}
\end{llmprompt}

\vspace{-0.5\baselineskip}

\begin{llmprompt}{Prompt For Only the Questions (Q2)}
    \#\#\# Question: \{question\}
    \bigskip

    \#\#\# Answer: \trainingTarget{\{answer\}}
\end{llmprompt}

\vspace{-0.5\baselineskip}

\begin{llmprompt}{Prompt For Only the Time Series Encoded as Text (Q3)}
    \#\#\# Instruction: You are provided with a 3-dimensional dataset of shape [320, 24, 3], representing 3-dimensional spatial locations of 24 human-body joints over 320 time steps. Using this data, please analyze the movements and provide the answer associated with this human activity sequence.
    \bigskip

    \#\#\# Data: \{scaled time series data\}
    \bigskip

    \#\#\# Answer: \trainingTarget{\{answer\}}
\end{llmprompt}

\vspace{-0.5\baselineskip}

\subsubsection{Prompts for \Cref{sec:bench:baselines}.}
The experiments \enquote{Ablation: Only Question} and \enquote{Ablation: Only TS} use the prompts in (Q2) and (Q3), respectively. The Naive Baseline uses the following:

\vspace{-0.5\baselineskip}

\begin{llmprompt}{Prompt For Naive Baseline (Time Series Encoded as Text and Questions)}
    \#\#\# Instruction: You are provided with a 3-dimensional dataset of shape [320, 24, 3], representing 3-dimensional spatial locations of 24 human-body joints over 320 time steps. Using this data, please analyze the movements and respond to the following question related to human activity recognition.
    \bigskip

    \#\#\# Data: \{scaled time series data\}
    \bigskip

    \#\#\# Question: \{question\}
    \bigskip

    \#\#\# Answer: \trainingTarget{\{answer\}}
\end{llmprompt}

\subsubsection{Measuring Language Modeling Deterioration}
\label{sec:app:eval-details:lm_retention}

For \Cref{tab:language-model-deterioration}, we evaluate the unchanged Llama~3.1~8B base model and all nine final adapters on the raw WikiText-2 test corpus from \texttt{Salesforce/wikitext}~\citep{merity2016pointer}.
We preserve the released row order, including empty rows, and concatenate rows with two newline characters.
All models use the same base tokenizer, without a task prompt, chat template, or additional special tokens.

We use overlapping windows of 2,048 tokens with a stride of 1,024 over the complete token stream.
Tokens already scored in an earlier window supply context only; every target token is scored exactly once, apart from the first corpus token, which has no preceding context.
This gives $N=289{,}075$ scored targets for every model and precision condition.
Models run in evaluation mode without gradients, with token cross-entropy computed in FP32.
We accumulate the summed negative log-likelihood before calculating
\[
    \overline{\ell}=\frac{1}{N}\sum_{i\in\mathcal{S}}-\log p_\theta(w_i\mid c_i),
    \qquad \mathrm{PPL}=\exp(\overline{\ell}),
\]
where $\mathcal{S}$ is the set of scored target positions and $c_i$ is the preceding context available within the corresponding window.
The ratio to the precision-matched base is $\exp(\overline{\ell}_{\mathrm{adapter}}-\overline{\ell}_{\mathrm{base}})$; it is a multiplicative factor, not a percentage.

The primary comparison loads the base weights in BF16 and attaches the adapters without merging them.
The additional NF4 comparison uses double quantization and BF16 computation, matching the training quantization configuration.

We also perform a small qualitative continuation experiment on five fixed 128-token Wikipedia prefixes for each tested model, generating greedily at most 24 new tokens.
The tested models are the NF4 base, question-only binary, time-series-only multiple-choice, Naive binary, and Naive open adapters.
The selected closed-answer adapters produce task-label repetitions or immediate termination; the base continues prose, and the Naive open adapter produces short sentence-like continuations that terminate early.

\begin{table}[h]
    \centering
    \caption{\textbf{Language modeling deteriorates drastically after fine-tuning for QuAnTS.}}
    \label{tab:language-model-deterioration}
    \denseColumns
    \begin{tabular}{lcrrrr}
        \toprule
        \multirow{2}{*}{System} & \multirow{2}{*}{\makecell[c]{Answer\\Type}} & \multicolumn{2}{c}{BF16} & \multicolumn{2}{c}{NF4} \\
        \cmidrule(lr){3-4}\cmidrule(lr){5-6}
         & & PPL (\textdownarrow) & PPL/base & PPL (\textdownarrow) & PPL/base \\
        \midrule
        Base & --- & 5.61 & 1.00 & 5.97 & 1.00 \\
        \subtleCmidrule{\cmidrule{1-6}}
        \multirow{3}{*}{\makecell[l]{Ablation:\\Only Question}} & Binary & $3.17\times10^{8}$ & $5.65\times10^{7}$ & $3.71\times10^{8}$ & $6.21\times10^{7}$ \\
         & Multi & 9870.98 & 1758.13 & $6.46\times10^{4}$ & $1.08\times10^{4}$ \\
         & Open & 19.24 & 3.43 & 25.65 & 4.30 \\
        \subtleCmidrule{\cmidrule{1-6}}
        \multirow{3}{*}{\makecell[l]{Ablation:\\Only TS}} & Binary & $2.10\times10^{7}$ & $3.74\times10^{6}$ & $1.74\times10^{7}$ & $2.92\times10^{6}$ \\
         & Multi & $6.17\times10^{8}$ & $1.10\times10^{8}$ & $7.32\times10^{8}$ & $1.23\times10^{8}$ \\
         & Open & 10.39 & 1.85 & 12.60 & 2.11 \\
        \subtleCmidrule{\cmidrule{1-6}}
        \multirow{3}{*}{\makecell[l]{Naive:\\TS + Question}} & Binary & $1.14\times10^{8}$ & $2.03\times10^{7}$ & $1.48\times10^{8}$ & $2.48\times10^{7}$ \\
         & Multi & $1.35\times10^{7}$ & $2.40\times10^{6}$ & $2.90\times10^{7}$ & $4.86\times10^{6}$ \\
         & Open & 10.61 & 1.89 & 12.99 & 2.18 \\
        \bottomrule
    \end{tabular}
\end{table}

\subsection{Measuring Human Performance}
\label{sec:app:eval-details:human}

\begin{figure}[p]
    \centering
    \includegraphics[width=0.75\linewidth]{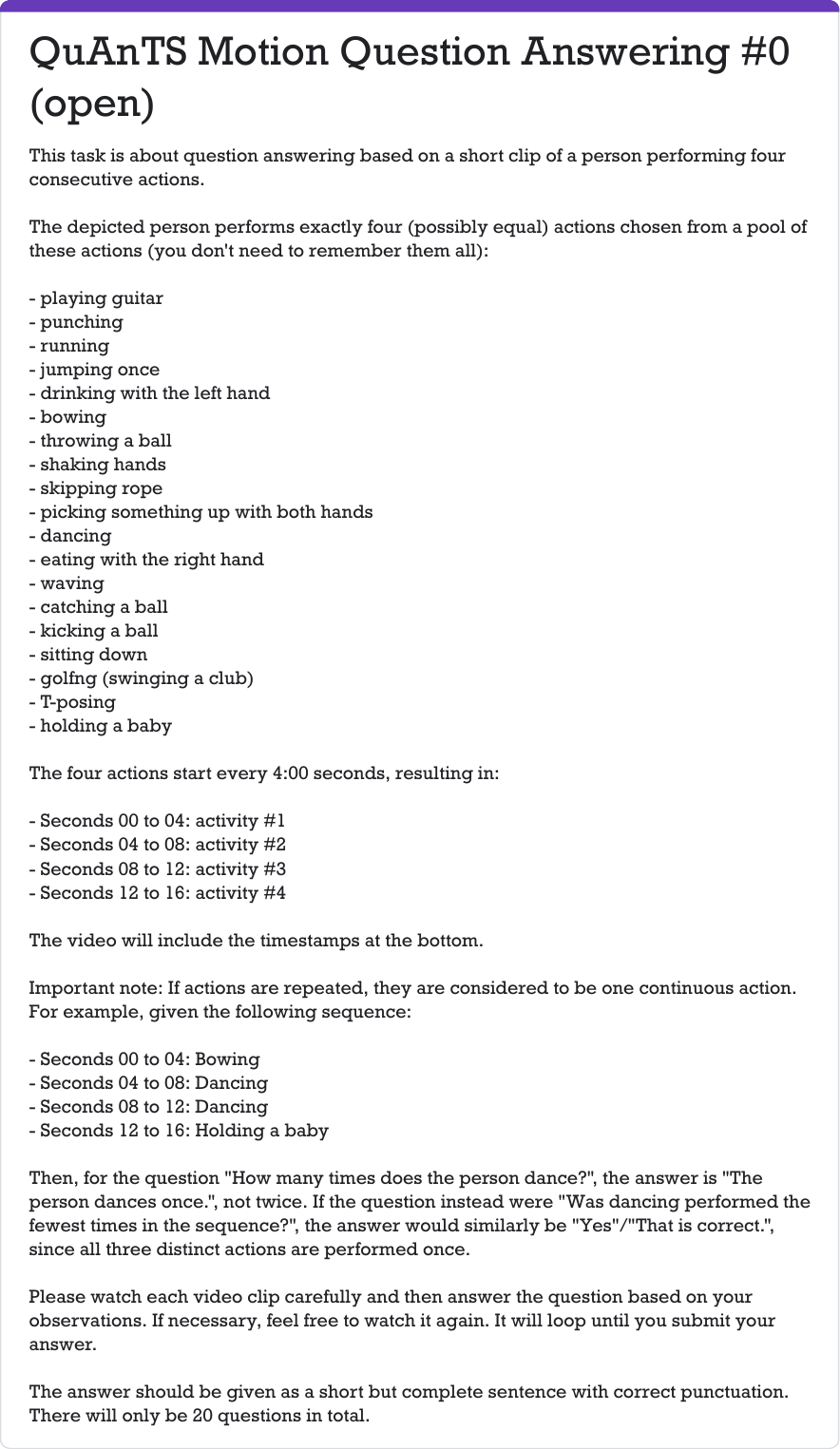}
    \caption{\textbf{Study participants were first presented with this introduction screen.}
        The text for the binary and multi variants only differed in the description of the answer modalities in the last three paragraphs.}
    \label{fig:study-excerpt:intro}
\end{figure}

\begin{figure}[p]
    \centering
    \begin{minipage}[b]{0.48\textwidth}
        \centering
        \begin{subfigure}[b]{\textwidth}
            \centering
            \includegraphics[width=\textwidth]{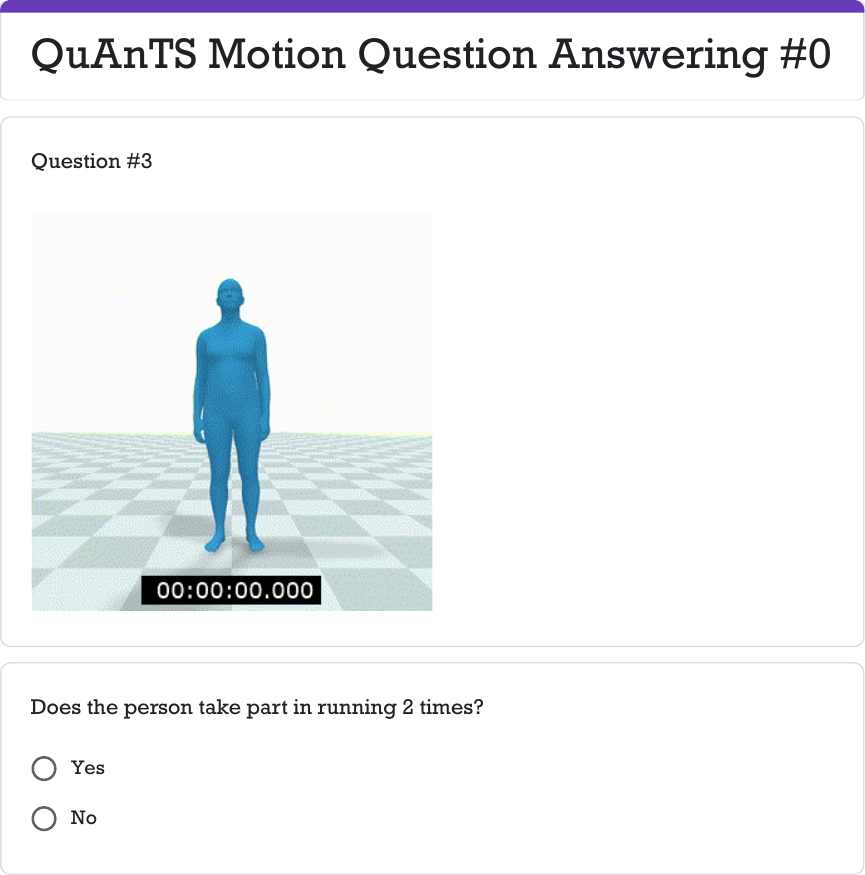}
            \caption{Binary}
            \label{fig:study-excerpt:questions:binary}
            \vspace{0.4cm}
        \end{subfigure}
        \begin{subfigure}[b]{\textwidth}
            \centering
            \includegraphics[width=\textwidth]{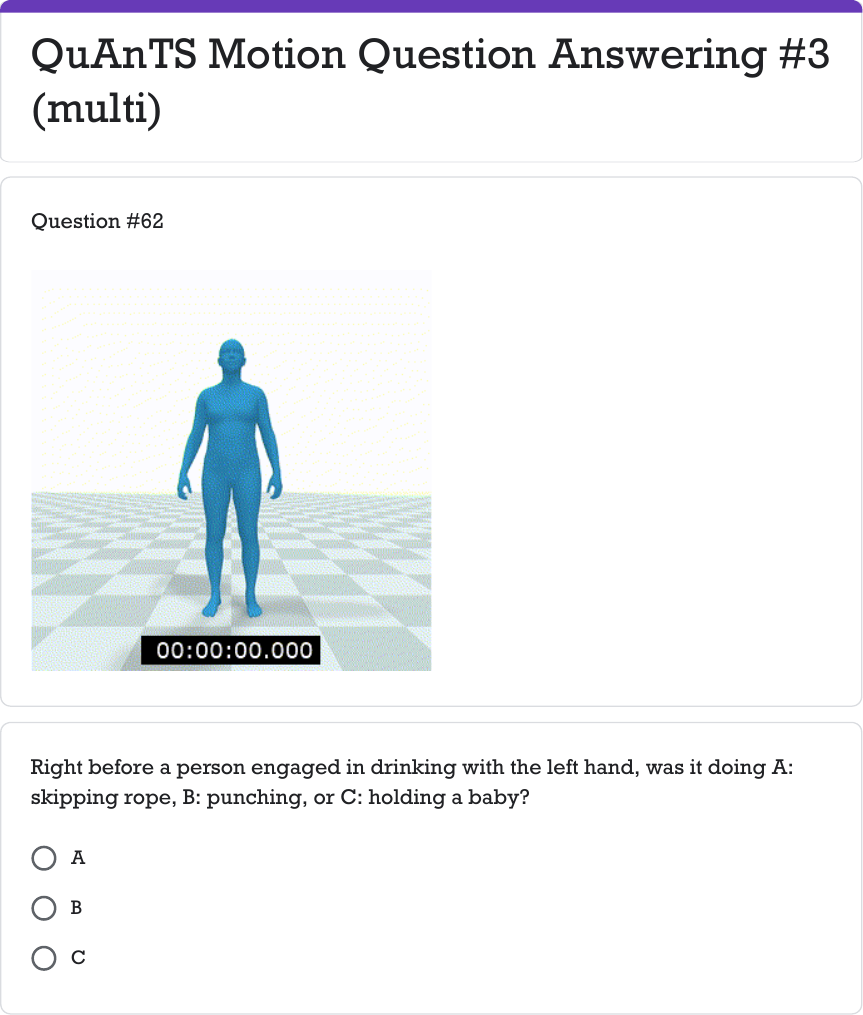}
            \caption{Multi}
            \label{fig:study-excerpt:questions:multi}
        \end{subfigure}
    \end{minipage}
    \hfill
    \begin{minipage}[b]{0.48\textwidth}
        \centering
        \begin{subfigure}[b]{\textwidth}
            \centering
            \includegraphics[width=\textwidth]{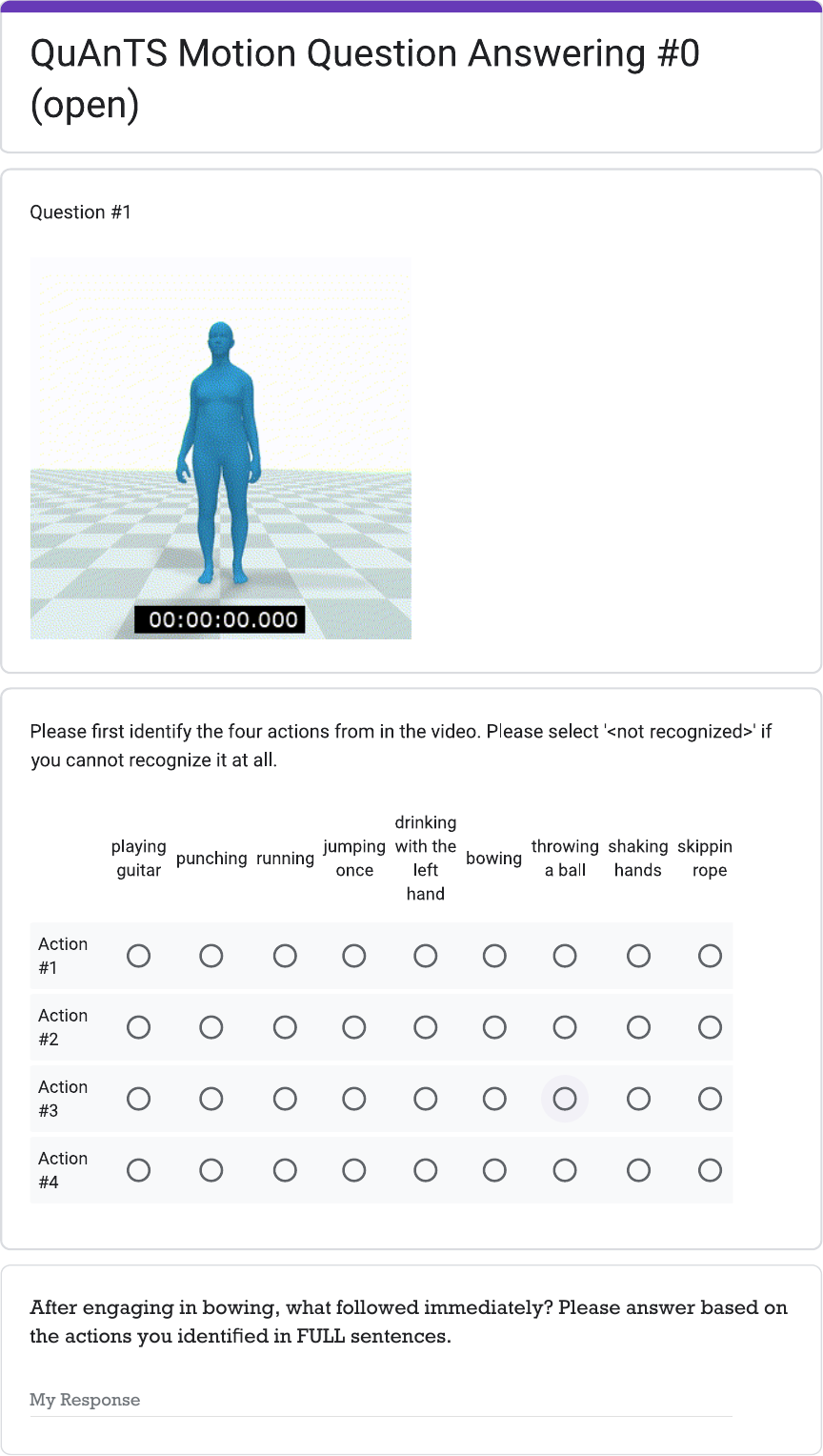}
            \caption{Open}
            \label{fig:study-excerpt:questions:open}
        \end{subfigure}
    \end{minipage}
    \caption{\textbf{Study participants were presented with screens matching the respective answer modality.}
        For the open variant, we also collected information on the identifiability of the action sequences.}
    \label{fig:study-excerpt:questions}
\end{figure}

\paragraph{Setup.}
To ensure clear stimuli, each page of the online study form showed the single relevant time series context as a rendered video clip using the gender-neutral SMPL-H model~\citep{loperSMPLSkinnedMultiperson2015,romeroEmbodiedHandsModeling2017}, and below the question with matching answer modality.
Participants were able to watch the context video clip as many times as they wished, even after seeing the question~(and possibly answer options).
The excerpts of the survey introduction screen in \Cref{fig:study-excerpt:intro} and questions in \Cref{fig:study-excerpt:questions} illustrate this.

\begin{figure}[h]
    \centering
    \includegraphics[width=\linewidth]{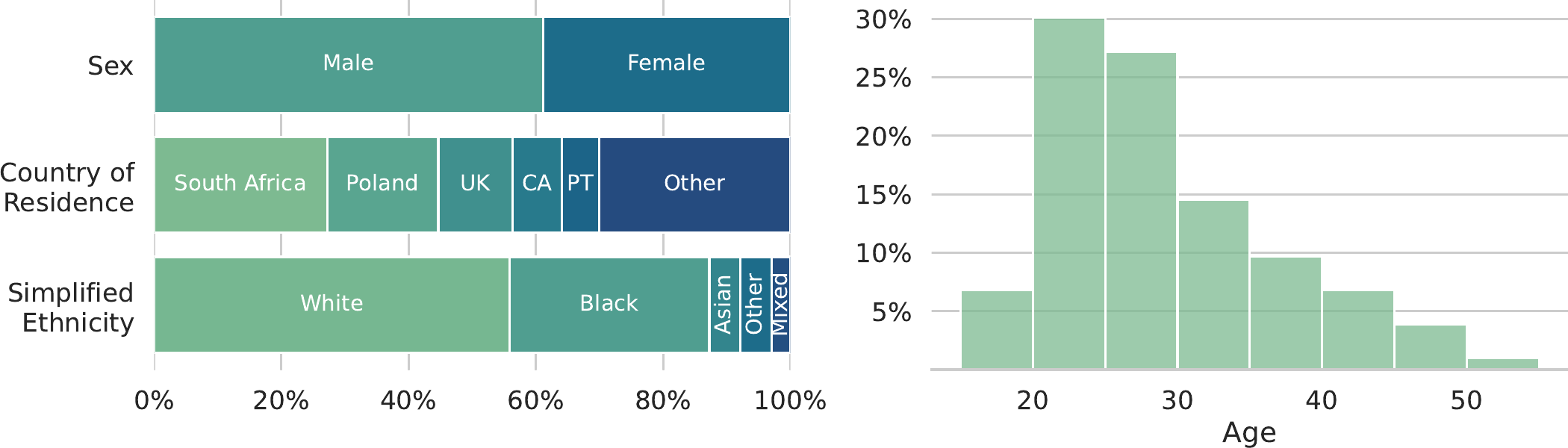}
    \caption{\textbf{The study was performed on a diverse set of 105 participants}, where each provided 20 answers spread over the three answer types.}
    \label{fig:human_eval-demographics}
    \vspace{-0.5\baselineskip}
\end{figure}

\paragraph{Demographics.}
We paid a fair median hourly wage of at least 12 Euros.
The diverse self-reported demographics are shown in \Cref{fig:human_eval-demographics}.

\begin{figure}[h]
    \centering
    \includegraphics[width=0.9\linewidth]{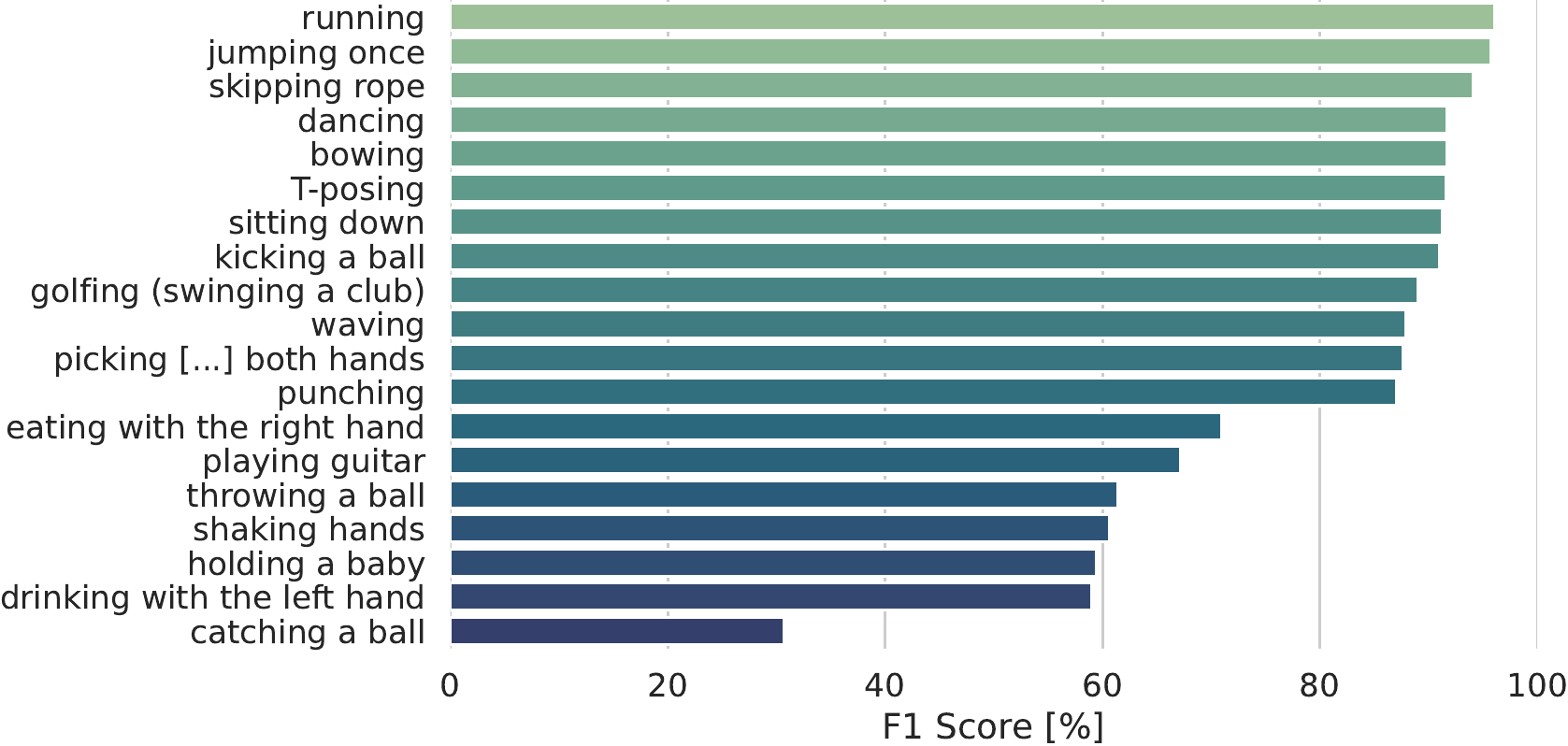}
    \caption{\textbf{Most actions are easily recognizable from the animated video clips.}}
    \label{fig:identifiability_actions}
    \vspace{-\baselineskip}
\end{figure}

\paragraph{Identification of Actions.}
In addition to the findings presented in the main body of the work, we evaluated how well humans can recognize individual actions within the sequences.
The data was obtained via the questions shown in \Cref{fig:study-excerpt:questions:open}.
For each of the $n = 440$ results from \Cref{tab:main-results}, four actions had to be identified.
For only 136~(7.73\%) cases out of those $440 \times 4 = 1,760$, the option \texttt{<not recognized>} was selected.
\Cref{fig:identifiability_actions} shows the binary F1 score, where unrecognized actions were counted as wrong.

\clearpage

\subsection{Adapting ChatTS to \datasetName{}}
\label{sec:app:eval-details:chatts}

Most importantly, to stay within the upper limit of the 30 variates that ChatTS can handle, we reduce the joints to only
\texttt{pelvis},
\texttt{left\_foot},
\texttt{right\_foot},
\texttt{head},
\texttt{left\_shoulder},
\texttt{right\_shoulder},
\texttt{left\_elbow},
\texttt{right\_elbow},
\texttt{left\_wrist}, and
\texttt{right\_wrist}, which we deemed the most discriminative for the actions in \datasetName{}.

Due to implementation limitations, we were unable to strictly constrain the tokens generated by ChatTS to adhere to the described format, resulting in slightly elevated rates of non-parsable responses.

We employed the following prompt:
\bigskip

\begin{llmprompt}{Prompt For ChatTS}
    <|im\_start|>system\\
    You are a helpful assistant.<|im\_end|><|im\_start|>user\\
    I have 30 time series representing human motion data from skeleton tracking. All time series are of length 320.\\
    The time series:
    \bigskip

    - pelvis x axis: <ts><ts/>\\{}
    [\dots]\\
    - right\_wrist z axis: <ts><ts/>
    \bigskip

    The recorded time series tracks four subsequent actions. Each takes 80 steps, reflecting four seconds of wall-clock time.
    \bigskip

    The possible action categories being performed are:\\
    - playing guitar\\{}
    [\dots]\\
    - holding a baby
    \bigskip

    First, identify the sequence of actions. If the same action occurs multiple times directly after one another, it is counted as a single continuous one. Only merge actions that have identical labels. Recall that there are four actions in total.
    \bigskip

    Finally, answer the following question in a full sentence: "\{question\}".
    \bigskip

    Only output the following:\\
    1. "Actions: <4 identified action names before merging>",\\
    2. "Merged Actions: <1 to 4 potentially merged actions>", and\\
    3. then the final response as "Answer: <your response>".<|im\_end|><|im\_start|>assistant\\
    Actions: 
\end{llmprompt}

\clearpage
\subsection{Details on xQA}
\label{sec:app:eval-details:nesy}

To evaluate the difficulty of \datasetName{}, we evaluate the xQA baseline family based on xLSTM-Mixer in \Cref{sec:bench:baselines}.
It factorizes the problem into (i) action extraction from the time series and (ii) question answering with an LLM.
We report results on two LLM backends, xQA--Qwen and xQA--LLaMA, with an otherwise identical pipeline.
To separate LLM reasoning quality from perception errors, we provide results on ground-truth action sequences~(GT) from the debugging metadata of \datasetName{} and the actual sequences predicted by our action encoder~(AE).

\begin{figure}[htp]
    \centering
    \includegraphics[width=\linewidth]{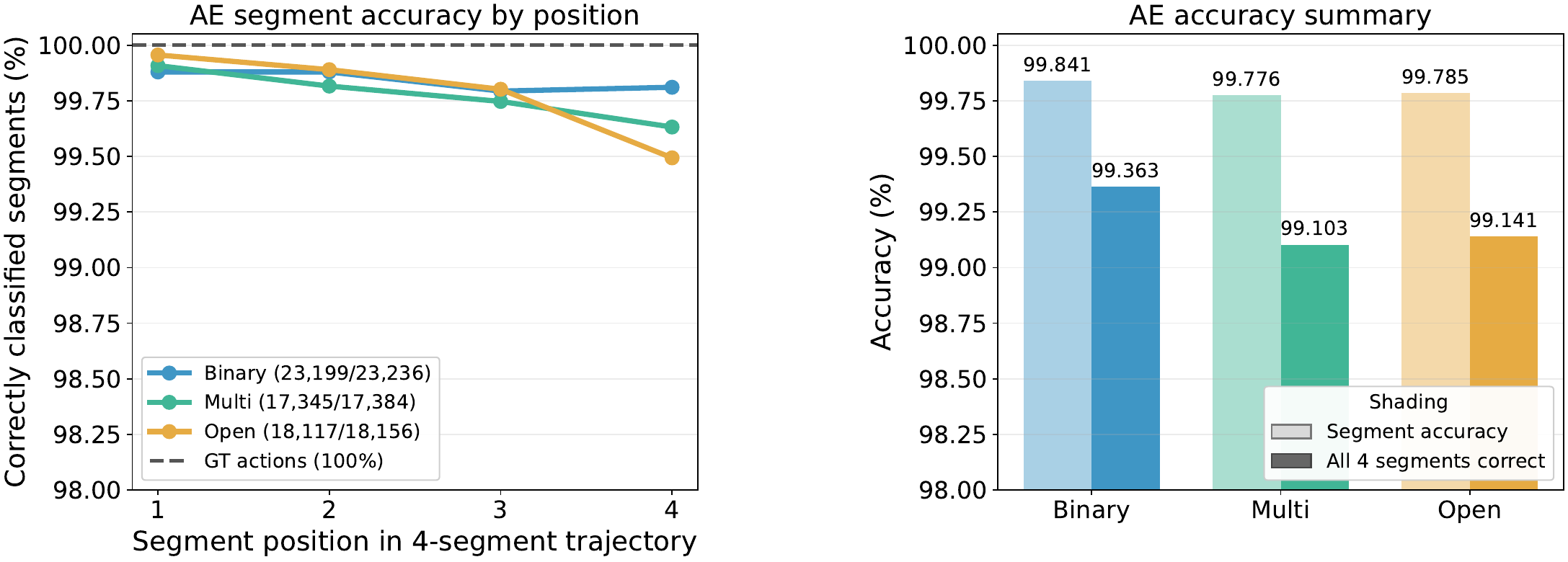}
    \caption{\textbf{The action encoder recognizes the 19 generated actions almost perfectly.} Visualized are the accuracies on the QuAnTS test set.}
    \label{fig:action_encoder_performance}
\end{figure}

\paragraph{Action Encoder.}
We utilize a segmentation model based on xLSTM-Mixer~\citep{kraus2025xlstmmixer}.
It is trained on the \datasetName{} action labels, which are not part of the regular data available to TSQA models.
The model predicts one of 19 actions per segment, i.e., using a human motion time series of length 80  with 72 variates.
The hidden size in the xLSTM blocks was set to 256, and training was performed using RAdam~\citep{liu2019variance} with a learning rate of $10^{-4}$.
We built the implementation on PyTorch~2.8 \citep{paszkePyTorchImperativeStyle2019} with Lightning~2.4 \citep{Falcon_PyTorch_Lightning_2019}.
For the xLSTM-Mixer implementation, we rely on a custom fork\footnote{\url{https://github.com/mauricekraus/transformers\#xlstm-mixer}} of the transformers library~\citep{wolfTransformersStateoftheArtNatural2020}.
The action encoder reached high accuracies above 99\% as detailed by \Cref{fig:action_encoder_performance}.

\paragraph{Prompt and Constrained Generation.}
Similar to the design of the LLMJudge laid out in \Cref{sec:app:llmjudge}, we explain the format in the prompt and constrain generated tokens to adhere strictly to the described JSON format using SGLang~\citep{zhengSGLangEfficientExecution2024}.
Both are prompted to return a single JSON object with fields \texttt{actions}, \texttt{steps}, and \texttt{answer}. The answer field is task-specific: for binary it must be \texttt{"true"}/\texttt{"false"}, for multi-choice it must be \texttt{"A"}/\texttt{"B"}/\texttt{"C"}, and for open-ended it is free-form. Invalid generations are retried a small number of times.
We used the following prompts for the three different answer types:

\begin{llmprompt}{Binary xQA Prompt}
    <|im\_start|>system\\
    You are a helpful timeseries question answering model.\\
    You are given a timeseries (TS) of **action names** and a question about them.\\
    If an action appears multiple times in a row in the TS, it counts as **one** occurrence (consecutive repeats collapse conceptually).\\
    Translate the timeseries into reasoning steps and then answer the question. Provide your thought process in "steps: [...]".
    \bigskip

    \#\# Example 1\\
    TS: ["running", "bowing", "jumping once", "playing guitar"], QS: Do the person's actions stay the same before and after they are jumping once?
    \begin{lstlisting}
{
    "actions": ["running", "bowing", "jumping once", "playing guitar"],
    "steps": [
        "1. Identify the index of \"jumping once\" which is 2.",
        "2. Now we need to look before and after index 2, thus index 1 and index 3.",
        "3. At actions[1] the action is bowing and actions[3] the action is playing guitar.",
        "4. Because these two actions are different, the answer is false."
    ],
    "answer": "false"
}
    \end{lstlisting}
    \bigskip

    \#\# Example 2\\
    TS: ["catching a ball", "running", "catching a ball", "running"], QS: Is the person conducting catching a ball the same amount as running?
    \begin{lstlisting}
{
    "actions": ["catching a ball", "running", "catching a ball", "running" ],
    "steps": [
        "1. Let's count catching a ball. Catching a ball occurs at index 0 and 2, thus 2 times.",
        "2. Let's count running. Running occurs at index 1 and 3, thus 2 times.",
        "3. Therefore, the answer is true."
    ],
    "answer": "true"
}
    \end{lstlisting}
    <|im\_end|><|im\_start|>user\\
    TS: ["\{action\_1\}", "\{action\_2\}", ..., "\{action\_N\}"], QS: \{question\}\\
    Respond with exactly one JSON object with the keys "actions", "steps", and "answer".\\
    <|im\_end|><|im\_start|>assistant
\end{llmprompt}

\begin{llmprompt}{Multiple-Choice xQA Prompt}
    <|im\_start|>system\\
    \textcolor{gray}{\textit{Initial Instruction as in the Binary xQA Prompt}}
    \bigskip

    \#\# Example 1\\
    TS: ["running", "jumping once", "shaking hands", "jumping once"], QS: Which activity does the person carry out 1 time? A: jumping once, B: shaking hands, C: golfing (swinging a club)
    \begin{lstlisting}
{
    "actions": ["running", "jumping once", "shaking hands", "jumping once"],
    "steps": [
        "1. Identify the actions that are performed only once. From the timeseries we see that running and shaking hands occur only once.",
        "2. Among A,B,C, only shaking hands matches this (so B).",
        "3. Therefore, the answer is **B**."
    ],
    "answer": "B"
}
    \end{lstlisting}
    \bigskip

    \#\# Example 2\\
    TS: ["golfing (swinging a club)", "golfing (swinging a club)", "shaking hands", "running"], QS: Which activity does the person carry out 0 times? A: golfing (swinging a club), B: shaking hands, C: punching
    \begin{lstlisting}
{
    "actions": ["golfing (swinging a club)", "golfing (swinging a club)", "shaking hands", "running"],
    "steps": [
        "1. Rule out actions performed > 0 times.",
        "2. golfing (swinging a club) appears (consecutive repeats count as 1), so rule out A.",
        "3. shaking hands appears once, so rule out B.",
        "4. punching does not appear, so answer is **C**."
    ],
    "answer": "C"
}
    \end{lstlisting}
    **Respond with exactly one JSON object and no additional explanation or narrative. If none of the options seem to be valid choose a random one.**\\
    <|im\_end|><|im\_start|>user\\
    TS: ["\{action\_1\}", "\{action\_2\}", ..., "\{action\_N\}"], QS: \{question with options A/B/C\}\\
    Respond with exactly one JSON object with the keys "actions", "steps", and "answer", where "answer" is one of "A", "B", or "C".\\
    <|im\_end|><|im\_start|>assistant
\end{llmprompt}

\begin{llmprompt}{Open-Ended xQA Prompt}
    <|im\_start|>system\\
    \textcolor{gray}{\textit{Initial Instruction as in the Binary xQA Prompt}}
    \bigskip

    \#\# Example 1\\
    TS: ["running", "punching", "punching", "playing guitar"], QS: Describe in plain English what the person is doing.
    \begin{lstlisting}
{
    "actions": ["running", "punching", "punching", "playing guitar"],
    "steps": [
        "1. Notice two consecutive occurrences of the same action count as one.",
        "2. Unique action sequence = [running, punching, playing guitar].",
        "3. Summarize in plain English."
    ],
    "answer": "The person runs, then punches, then plays guitar."
}
    \end{lstlisting}
    \bigskip

    \#\# Example 2\\
    TS: ["jumping once", "jumping once", "shaking hands", "golfing (swinging a club)", "golfing (swinging a club)", "golfing (swinging a club)", "holding a baby"], QS: Explain step by step how many times each action occurs, and then summarize.
    \begin{lstlisting}[mathescape]
{
    "actions": ["jumping once", "jumping once", "shaking hands", "golfing (swinging a club)", "golfing (swinging a club)", "golfing (swinging a club)", "holding a baby"],
    "steps": [
        "1. Count unique consecutive:",
        "   - jumping once appears consecutively $\Rightarrow$ count=1",
        "   - shaking hands appears once $\Rightarrow$ count=1",
        "   - golfing (swinging a club) appears consecutively $\Rightarrow$ count=1",
        "   - holding a baby appears once $\Rightarrow$ count=1",
        "3. Totals: jumping once=1, shaking hands=1, golfing (swinging a club)=1, holding a baby=1.",
        "4. Summarize in one sentence."
    ],
    "answer": "The person jumps once, shakes hands, plays golf with a club, and then holds a baby."
}
    \end{lstlisting}
    **Respond with exactly one JSON object and no additional explanation or narrative.**\\
    <|im\_end|><|im\_start|>user\\
    TS: ["\{action\_1\}", "\{action\_2\}", ..., "\{action\_N\}"], QS: \{free-form question\}\\
    Respond with exactly one JSON object with the keys "actions", "steps", and "answer".\\
    <|im\_end|><|im\_start|>assistant
\end{llmprompt}

\FloatBarrier

\section{Extended Results}
\label{sec:app:results}

\subsection{Action Recognizability and Human Performance}
\label{sec:app:results:human_detection_and_perf}

\begin{figure}[htp]
    \centering
    \vspace{\baselineskip}
    \includegraphics[width=\linewidth]{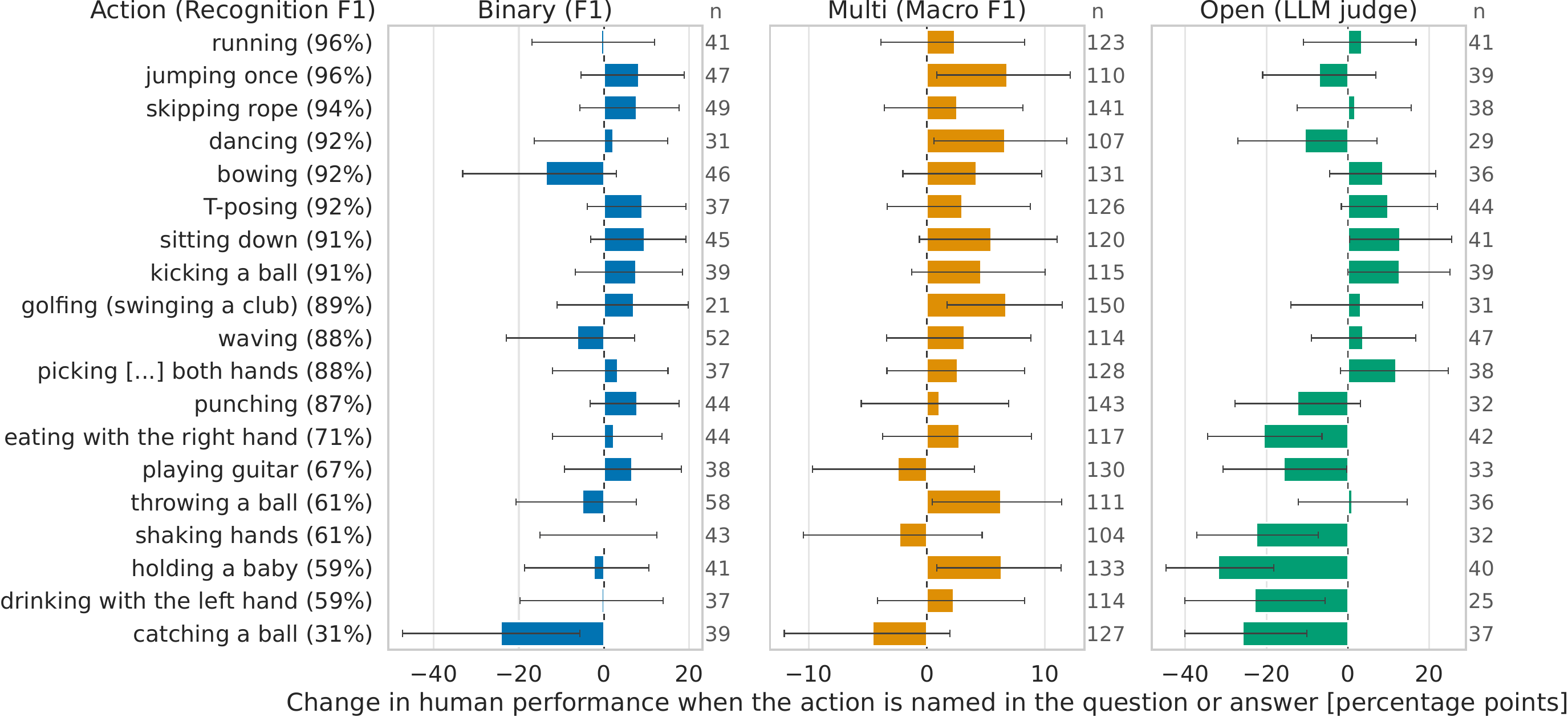}
    \caption{\textbf{The presence of different actions in the question or answer drives changes in QA performance.}
    For open, the relation is less obvious, since in 2 out of 3 cases, a hard-to-detect action is in a distractor rather than the correct answer.}
    \label{fig:action_presence}
\end{figure}

\begin{figure}[htp]
    \centering
    \includegraphics[width=\linewidth]{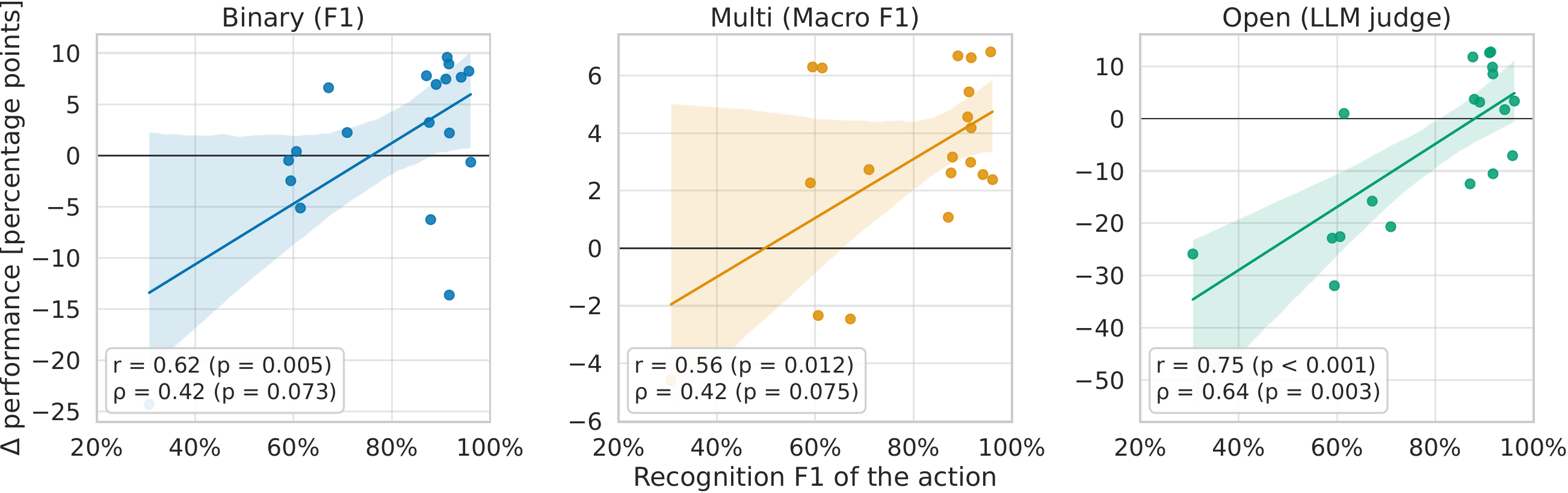}
    \caption{\textbf{Overall, action recognizability is positively correlated with performance.}
    We also report the respective Person's correlation coefficient $r$ and Spearman's rank correlation coefficient $\rho$, along with their $p$-values.
    The relationship is significant only for the open split.}
    \label{fig:action_presence_correlation_perf}
\end{figure}

We also assess whether human performance depends on the humans' ability to recognize actions in the provided video clips, or on linguistic and reasoning abilities.
\Cref{fig:action_presence} shows that the presence of certain action labels in the question or answer has an influence on the respective task performance.
This positive correlation is explored quantitatively in \Cref{fig:action_presence_correlation_perf}, which shows that the relationship is particularly pronounced for the open split.
This partly explains the shortcomings of human responses to certain actions~(see \Cref{sec:bench:humans}).

\subsection{Results by Answer and Question Type}
\label{sec:app:results:details}

\begin{figure}[h]
    \centering
    \includegraphics[width=0.95\linewidth]{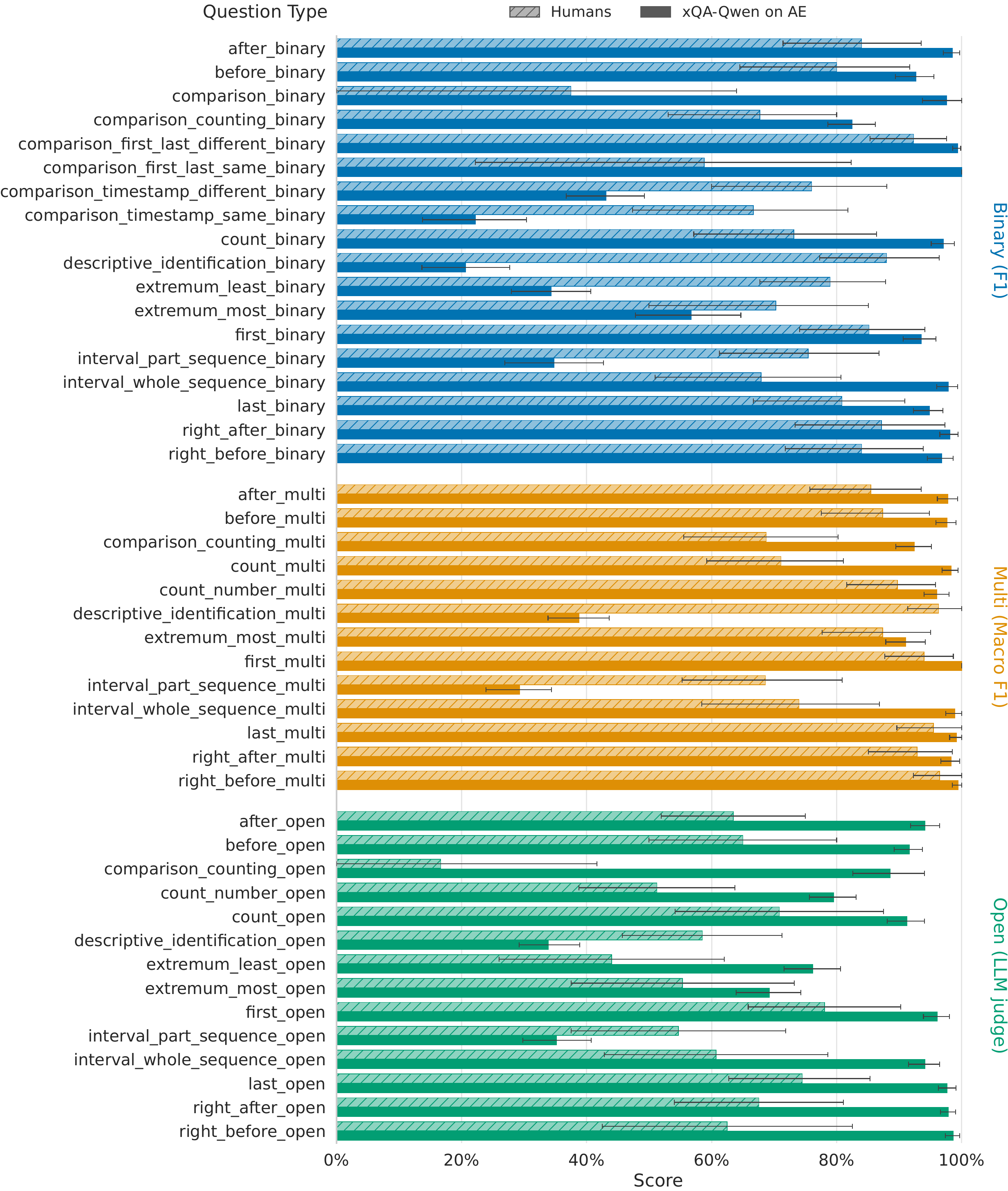}
    \caption{\textbf{xQA-Qwen on AE is ahead of humans on 35 of the 45 question types, especially on the multi and open splits.}
    The clearest exception to that is descriptive\_identification, where humans lead decisively across splits.
    The performance of humans ($n = 2{,}100$ responses) and xQA (all $n = 15{,}000$ pairs) is shown with 95-percentile bootstrap intervals from $2{,}000$ resamples, with wider ones for humans due to fewer samples.}
    \label{fig:app:detailed_results}
\end{figure}

A more detailed breakdown of selected results is provided in \Cref{fig:app:detailed_results}.

\clearpage
\section{Datasheet for \datasetName{}}
\label{sec:app:datasheet}

In compliance with the DMLR Submission Guidelines\footnote{See \url{https://data.mlr.press/submissions}.}, %
this section provides a datasheet following \citet{gebruDatasheetsDatasets2021}.
Specifically, it contains documentation and intended uses~(\ref{sec:app:datasheet:uses}), links for viewing and obtaining the dataset~(\ref{sec:app:datasheet:distribution}), as well as a hosting, licensing, and maintenance plan~(\ref{sec:app:datasheet:distribution} \& \ref{sec:app:datasheet:maintenance}).

\newcommand{\dsection}[1]{\subsection{#1}}
\newcommand{\dquestion}[1]{\paragraph*{#1}}
\newcommand{\danswer}[1]{$\vartriangleright$~\textit{#1}}

\dsection{Motivation}
\label{sec:app:datasheet:motivation}

\dquestion{For what purpose was the dataset created?}
\danswer{\datasetName{} provides both training and evaluation data for time series question answering~(TSQA). It poses textual questions regarding numerical multivariate time series and provides reference answers alongside additional metadata.}

\dquestion{Who created the dataset and on behalf of which entity?}
\danswer{The dataset was created by Felix Divo and Maurice Kraus at the AIML Group, Department of Computer Science, Technische Universität Darmstadt in Darmstadt, Germany.}

\dquestion{Who funded the creation of the dataset?}
\danswer{The work was funded by multiple entities as listed in the Section \enquote{Acknowledgments and Disclosure of Funding} following \Cref{sec:conclusion}.}

\dsection{Composition}
\label{sec:app:datasheet:composition}

\dquestion{What do the instances that comprise the dataset represent?}
\danswer{The dataset comprises triplets of (1) a numerical time series of synthetic human motion tracking data of four actions, (2) a textual question, and (3) an answer. The latter can be binary (Yes/No), multiple-choice (one of A/B/C), or open (free text).
In addition, \datasetName{} provides supplemental textual versions for all answer types, the underlying action sequences (for debugging only), ground-truth scene descriptions (for debugging only), and the question/answer type for statistical purposes.}

\dquestion{How many instances are there in total (of each type, if appropriate)?}
\danswer{We generated 30.000 time series contexts, each with five question-answer pairs, resulting in a total of 150.000 instances.}

\dquestion{Does the dataset contain all possible instances, or is it a sample of instances from a larger set?}
\danswer{\datasetName{} was created synthetically from scratch to reach a set target size.
We put in great effort to ensure diversity of questions and answers~(see, in particular, \Cref{sec:dataset:stats}) and a balanced distribution of question and answer types~(see \Cref{sec:app:details-generate-dataset}).
The probabilistic motion diffusion process ensures sufficient diversity of the numerical time series.}

\dquestion{What data does each instance consist of? \enquote{Raw} data (e.g., unprocessed text or images) or features?}
\danswer{The components of each instance are described above.
Each time series is a motion trajectory of 16 seconds at 20 frames per second~(320 steps) of 24 joints with a 3-dimensional position, and thus a tensor of rank $320 \times 24 \times 3$.}

\dquestion{Is there a label or target associated with each instance?}
\danswer{The target value to predict is the answer, which is in either binary, multiple-choice, or open format.}

\dquestion{Is any information missing from individual instances?}
\danswer{No.}

\dquestion{Are relationships between individual instances made explicit?}
\danswer{The samples are independent and identically distributed~(i.i.d).}

\dquestion{Are there recommended data splits~(e.g., training, development/validation, testing)?}
\danswer{For best comparability, we split the dataset into pre-defined training, validation, and testing splits of 120.000, 15.000, and 15.000 samples, respectively.}

\dquestion{Are there any errors, sources of noise, or redundancies in the dataset? If so, please provide a description.}
\danswer{The synthetic generation process limits potential sources of error.
Beyond manual inspection, we provide extensive validation and ablation experiments in \Cref{sec:bench:verify}.
The vast majority of question-answer pairs are unique, as shown in \Cref{tab:dataset-stats-comparison}.
Possible errors can arise if the human motion generation fails to faithfully represent the conditioning activity description.}

\dquestion{Is the dataset self-contained, or does it link to or otherwise rely on external resources?}
\danswer{It is fully self-contained.}

\dquestion{Does the dataset contain data that might be considered confidential?}
\danswer{No.}

\dquestion{Does the dataset contain data that, if viewed directly, might be offensive, insulting, threatening, or otherwise cause anxiety?}
\danswer{No.}

\dsection{Collection Process}
\label{sec:app:datasheet:collection}

\dquestion{How was the data associated with each instance acquired? What mechanisms or procedures were used to collect the data?}
\danswer{The dataset was created synthetically, as detailed in \Cref{sec:dataset}.
We sampled random actions and generated randomized matching human motions.
For each instance, a question-answer type was chosen randomly, for which we instantiate a randomly selected text template.}

\dquestion{If the dataset is a sample from a larger set, what was the sampling strategy?}
\danswer{n/a}

\dquestion{Who was involved in the data collection process (e.g., students, crowdworkers, contractors) and how were they compensated?}
\danswer{No third parties were involved in the dataset's creation.
Crowdworkers assisted in evaluating the dataset.
The details, including their fair compensation, are provided in \Cref{sec:bench:humans,sec:app:eval-details:human}.}

\dquestion{Over what timeframe was the data collected?}
\danswer{See above for context.
The evaluation data was collected in Q3 and Q4 of 2024.}

\dquestion{Were any ethical review processes conducted (e.g., by an institutional review board)?}
\danswer{No such review was deemed necessary as the study posed negligible risk: adult participants viewed synthetic, non-sensitive animations and answered comprehension questions, posing no risks beyond those of daily life.
Voluntary withdrawal was possible at any point and a fair compensation of at least 12 Euros per hour median was provided.
We report demographics only in aggregate~(see \Cref{sec:app:eval-details:human}, specifically \Cref{fig:human_eval-demographics}).}

\dsection{Preprocessing/cleaning/labeling}
\label{sec:app:datasheet:preprocessing}

\dquestion{Was any preprocessing/cleaning/labeling of the data done?}
\danswer{No, the desired target format was generated directly.}

\dquestion{Was the \enquote{raw} data saved in addition to the preprocessed/cleaned/labeled data?}
\danswer{n/a}

\dquestion{Is the software that was used to preprocess/clean/label the data available?}
\danswer{The software used to generate the entire dataset is available at \linkGeneration{}.}

\dsection{Uses}
\label{sec:app:datasheet:uses}

\dquestion{Has the dataset been used for any tasks already?}
\danswer{We provide initial experiments using the brand new dataset in \Cref{sec:bench}.}

\dquestion{Is there a repository that links to any or all papers or systems that use the dataset?}
\danswer{n/a}

\dquestion{What (other) tasks could the dataset be used for?}
\danswer{We anticipate usage to focus on TSQA.}

\dquestion{Is there anything about the composition of the dataset or the way it was collected and preprocessed/cleaned/labeled that might impact future uses? For example, is there anything that a dataset consumer might need to know to avoid uses that could result in unfair treatment of individuals or groups (e.g., stereotyping, quality of service issues) or other risks or harms (e.g., legal risks, financial harms)? Is there anything a dataset consumer could do to mitigate these risks or harms?}
\danswer{We anticipate very few of such issues.
Potential misrepresentation of actions could be inherited from STMC, e.g., as it does not represent people with certain (motoric) disabilities.
This is a very common limitation in human motion data.
Consumers of the dataset should be aware of these limitations when deploying systems in the real world.}

\dquestion{Are there tasks for which the dataset should not be used?}
\danswer{No (see previous).}

\dsection{Distribution}
\label{sec:app:datasheet:distribution}

\dquestion{Will the dataset be distributed to third parties outside of the entity on behalf of which the dataset was created?}
\danswer{Yes.}

\dquestion{How will the dataset be distributed? Does the dataset have a digital object identifier~(DOI)?}
\danswer{The dataset is freely available at \linkDataset{}, which provides various common data formats.
The DOI of the dataset is \doiDataset{}.}

\dquestion{Will the dataset be distributed under a copyright or other intellectual property license, and/or under applicable terms of use?}
\danswer{The dataset is available under the permissive CC BY 4.0 license, see \url{https://creativecommons.org/licenses/by/4.0}.}

\dquestion{Have any third parties imposed IP-based or other restrictions on the data associated with the instances?}
\danswer{We are aware of none.}

\dquestion{Do any export controls or other regulatory restrictions apply to the dataset or to individual instances?}
\danswer{We are aware of none.}

\dsection{Maintenance}
\label{sec:app:datasheet:maintenance}

\dquestion{Who will be supporting/hosting/maintaining the dataset?}
\danswer{\datasetName{} is distributed by Huggingface (DOI~\doiDataset{}). We would seek other means of archival and distribution, such as university repositories, if long-term accessibility became uncertain at some point in the future.}

\dquestion{How can the owner/curator/manager of the dataset be contacted~(e.g., email address)?}
\danswer{Email adresses of all authors are provided on p.~\pageref{first_page}.}

\dquestion{Is there an erratum?}
\danswer{We will publish any important changes on the respective repositories for generation~(\linkGeneration{}), evaluation~(\linkGeneration{}), or \datasetName{} itself~(\linkDataset).}

\dquestion{Will the dataset be updated (e.g., to correct labeling errors, add new instances, or delete instances)?}
\danswer{We currently do not plan any updates.}

\dquestion{If the dataset relates to people, are there applicable limits on the retention of the data associated with the instances (e.g., were the individuals in question told that their data would be retained for a fixed period of time and then deleted)?}
\danswer{n/a}

\dquestion{Will older versions of the dataset continue to be supported/hosted/maintained?}
\danswer{Yes.}

\dquestion{If others want to extend/augment/build on/contribute to the dataset, is there a mechanism for them to do so?}
\danswer{The license permits freely adapting the dataset.
However, we currently do not plan to collect any community contributions.}

\end{document}